\documentclass{article}

 \usepackage[preprint]{neurips_2026}

\usepackage[utf8]{inputenc} 
\usepackage[T1]{fontenc}    
\usepackage{hyperref}       
\usepackage{url}            
\usepackage{booktabs}       
\usepackage{amsfonts}       
\usepackage{nicefrac}       
\usepackage{microtype}      
\usepackage{xcolor}         
\usepackage{booktabs}       
\usepackage{multirow}       
\usepackage{graphicx}       
\usepackage{amsmath}
\usepackage{amsthm}
\usepackage{algorithm}
\usepackage{algorithmic}
\newtheorem{remark}{Remark} 
\newtheorem{proposition}{Proposition}

\usepackage{subcaption} 
\usepackage{amssymb}
\usepackage{float} 
\hypersetup{hidelinks} 
\usepackage{blindtext}

\usepackage{amsmath,amsfonts,bm}

\def\eqref#1{equation~\ref{#1}}

\def\1{\bm{1}}

\DeclareMathAlphabet{\mathsfit}{\encodingdefault}{\sfdefault}{m}{sl}
\SetMathAlphabet{\mathsfit}{bold}{\encodingdefault}{\sfdefault}{bx}{n}

\newcommand{\M}{\mathbf}
\newcommand{\E}{\mathbb{E}}

\newcommand{\R}{\mathbb{R}}

\title{P-Flow: Proxy-gradient Flows for Linear Inverse Problems}

\author{%
  Zehua Jiang \thanks{Equal contribution.}\\
  Zhejiang University \\
  Hangzhou, China \\
  \texttt{jiangzehua@zju.edu.cn} \\
  \And
  Fenghao Zhu \footnotemark[1]\\
  Zhejiang University \\
  Hangzhou, China \\
  \texttt{zjuzfh@zju.edu.cn} \\
  \And
  Xinquan Wang \\
  Zhejiang University \\
  Hangzhou, China \\
  \texttt{wangxinquan@zju.edu.cn} \\
  \AND
  Chongwen Huang\\
  Zhejiang University \\
  Hangzhou, China \\
  \texttt{chongwenhuang@zju.edu.cn} \\
  \And
  Zhaoyang Zhang \\
  Zhejiang University \\
  Hangzhou, China \\
  \texttt{zhzy@zju.edu.cn} \\
}

\begin{document}

\maketitle

\begin{abstract}
Generative models based on flow matching have emerged as a powerful paradigm for inverse problems, offering straighter trajectories and faster sampling compared to diffusion models. However, existing approaches often necessitate differentiating through unrolled paths, leading to numerical instability and prohibitive computational overhead. To address this, we propose P-Flow, a framework that stabilizes the reconstruction process by leveraging a proxy gradient to update the source point. This approach effectively circumvents the numerical instability and memory overhead of long-chain differentiation. To ensure consistency with the prior distribution, we employ a Gaussian spherical projection motivated by the concentration of measure phenomenon in high-dimensional spaces. We further provide a theoretical analysis for P-Flow based on Bayesian theory and Lipschitz continuity. Experiments across diverse restoration tasks demonstrate that P-Flow delivers competitive performance, especially under extreme degradations such as severely ill-posed conditions and high measurement noise.
\end{abstract}

\section{Introduction}
Solving linear inverse problems is a central challenge in high-dimensional signal processing and machine learning. The goal is to reconstruct an underlying signal $\mathbf{x} \in \mathbb{R}^n$ from a degraded or incomplete observation $\mathbf{y} \in \mathbb{R}^m$. In the context of computational imaging, this task is widely known as image restoration and is typically formulated as:
\begin{equation}
    \mathbf{y} = \mathbf{H}\mathbf{x} + \mathbf{n},
\end{equation}
where $\mathbf{H} : \mathbb{R}^n \rightarrow \mathbb{R}^m$ with $m \leq n$ represents a linear operator and $\mathbf{n}$ denotes additive noise from an arbitrary distribution, which is typically modeled as white Gaussian noise $\mathcal{N}(0, \sigma^2 \mathbf{I})$. Since $\mathbf{H}$ is ill-conditioned or rank-deficient, the problem is inherently under-determined, requiring the prior knowledge about the natural image distribution $p_{\M X}$ to achieve a high-fidelity reconstruction. Traditionally, this challenge has been addressed by minimizing an objective function of the form:
\begin{equation}
\hat{\mathbf{x}} = \arg\min_{\mathbf{x}} \mathcal{L}(\mathbf{x}, \mathbf{y}) + \lambda \mathcal{R}(\mathbf{x}),
\end{equation}
where the loss $\mathcal{L}$ maintains data fidelity and $\mathcal{R}(\M x)$ is a regularization term of the prior, such as total variation \citep{rudin1992nonlinear} or wavelet sparsity \citep{donoho2006compressed}. From a Bayesian perspective, this minimization is equivalent to maximum a posteriori (MAP) estimation, derived from Bayes' rule as: 
\begin{equation}\label{eq:maximum A posteriori}
\hat{\mathbf{x}} = \arg\max_{\mathbf{x} \in \mathbb{R}^n} \log p(\mathbf{x}|\mathbf{y}) = \arg\min_{\mathbf{x} \in \mathbb{R}^n} { -\log p(\mathbf{y}|\mathbf{x}) - \log p(\mathbf{x}) },
\end{equation}
where $p(\mathbf{y}|\mathbf{x})$ is the likelihood determined by the degradation model and $p(\mathbf{x})$ denotes the prior distribution of natural images. 
\par
While early methods relied on handcrafted priors, the development of deep learning allowed supervised `task-specific' networks \citep{dong2015image, zhang2017beyond} to directly learn the mapping from $\mathbf{y}$ to $\hat{\M x}$. However, these models lack flexibility and often generalize poorly to inference-time degradations that exhibit a discrepancy with training-stage assumptions \citep{antun2020instabilities}. To overcome this, zero-shot approaches like Plug-and-Play (PnP) \citep{venkatakrishnan2013plug} and regularization by denoising (RED) \citep{romano2017little} emerged. These frameworks leverage pre-trained generative models as implicit image priors within iterative optimization frameworks, effectively bypassing the need for task-specific retraining.
\par
Leveraging generative priors, denoising diffusion probabilistic models (DDPMs) \citep{ho2020denoising, song2020score} serve as expressive priors by reversing a gradual Gaussian noising process and approximating the score function $\nabla_{\mathbf{x}} \log p(\mathbf{x})$. Methods like diffusion posterior sampling (DPS) \citep{chung2023diffusion} and pseudoinverse-guided diffusion Models ($\Pi$GDM) \citep{song2023pseudoinverse} incorporate gradient guidance to steer the diffusion sampling process toward measurement consistency. However, these methods often require a high number of function evaluations \citep{kawar2023denoising} and can be sensitive to the stochasticity of the diffusion path. To improve efficiency, flow matching (FM) \citep{lipman2023flow} offers a competitive paradigm. FM models learn a vector field $\mathbf{v}_t$ that defines an ordinary differential equation (ODE) trajectory between a source distribution and the data manifold. Using straighter probability paths, such as those derived from optimal transport (OT) \citep{lipman2023flow, pooladian2023multisample}, FM enables faster sampling and more stable deterministic trajectories. However, guiding these flows toward measurement consistency remains challenging \citep{pokle2023training}, as naive gradient corrections can pull samples off the data manifold. To address this, recent works integrate FM into PnP and Bayesian frameworks. PnP-Flow \citep{martin2025pnpflow} uses FM as a time-dependent denoiser to bypass costly ODE backpropagation. Simultaneously, Flower \citep{pourya2026flower} unifies PnP and generative solvers through a Bayesian ancestral-sampling lens, alternating between destination estimation and measurement refinement.
\par
These methodologies primarily leverage the posterior sampling guidance. An alternative paradigm, latent source optimization has shown promise for high-fidelity restoration by seeking the optimal latent source that generates the target reconstruction (D-Flow~\citep{ben2024d}). However, its implementation within the FM framework faces a significant hurdle. Specifically, source optimization requires differentiating through a long chain of neural network passes across the unrolled trajectory. The depth of this chain can induce gradient instability and demanding excessive memory and time. To address these challenges, we propose P-Flow, an efficient framework that enhances source optimization for FM. Instead of differentiating through complex ODE paths, P-Flow employs a proxy-gradient strategy to bypass unstable Jacobian chains and stabilize and accelerate convergence. Furthermore, we incorporate a Gaussian spherical projection to constrain the source point within the typical region of the prior. Our main contributions are summarized as follows:
\begin{enumerate}
    \item \textbf{P-Flow framework}: We introduce P-Flow, a novel framework for linear inverse problems that optimizes the source point via a three-step iterative process: (i) a time-consistent flow trajectory to generate the sample $\M x_1$ from latent source $\M x_0$; (ii) a gradient step on the data-fidelity loss that updates $\M x_0$ with a  proxy gradient; (iii) a Gaussian spherical projection to ensure the latent variables remain within the high-probability region of the prior.
    \item \textbf{Theoretical analysis and efficiency}: We provide a theoretical analysis for our proxy-gradient strategy based on Bayesian theory and Lipschitz continuity. P-Flow not only enhances optimization stability but also achieves a significant reduction in memory footprint and inference latency compared to previous source-optimization methods.
    \item \textbf{Numerical validation}: We conduct numerical evaluations on FFHQ and AFHQ-Cat. Results demonstrate that P-Flow achieves competitive performance across various restoration tasks, especially under extreme degradations such as severely ill-posed conditions and high measurement noise. 
\end{enumerate}

The remainder of this paper is organized as follows. Section \ref{sec:background} reviews the background of FM and the requisite mathematical foundations. Section \ref{sec:method} presents our main methodology along with the associated theoretical analysis. Section \ref{sec:related works} provides a comprehensive survey of related works. Section \ref{sec:numerical results} presents numerical results to demonstrate the effectiveness of our approach.

\section{Background}
\label{sec:background}
In this section, we provide the mathematical foundations for the proposed method and FM models. 

\subsection{Flow matching}
FM \cite{lipman2023flow} provides a framework for learning a time-dependent vector field $\M v_t^\theta: \R^d \to \R^d$ that generates a flow $\psi_t$. This flow defines a probability path $p_{\M X_t}$ that transforms a source distribution $p_{\M X_0}$ into the target data distribution $p_{\M X_1}$. The evolution of a sample $\M x$ along this trajectory is governed by the ODE:
\begin{equation} 
    \frac{\mathrm{d}}{\mathrm{d}t}\psi_t(\M x) = \M v_t\big(\psi_t(\M x)\big), \quad t \in [0,1].
\end{equation}
The ideal optimization objective is to minimize the difference between the parameterized velocity field $\M v_t^\theta$ and the true marginal velocity field $\M v_t$:
\begin{equation} 
\label{eq:fm_objective}
    \mathcal{L}_{\mathrm{FM}}(\theta) = \E_{t, \M x_t \sim p_{\M X_t}} \left[ \left\| \M v_t^{\theta}(\M x_t) - \M v_t(\M x_t) \right\|_2^2 \right].
\end{equation}
 However, directly computing $\M v_t$ is often intractable. To circumvent this, \cite{lipman2023flow} introduced the conditional flow matching (CFM) objective, which regresses $\M v_t^\theta$ against a simpler conditional velocity field $\M v_t(\M x_t \mid \M x_1)$. Assuming a straight-line flow between samples $\M x_0 \sim p_{\M X_0}$ and $\M x_1 \sim p_{\M X_1}$, we define:
\begin{equation}
    \M x_t = (1-t)\M x_0 + t\M x_1, \quad \M v_t(\M x_t \mid \M x_1) = \M x_1 - \M x_0.
\end{equation}
This leads to the practical training objective:
\begin{equation}
\label{eq:cfm_loss}
    \mathcal{L}_{\mathrm{CFM}}(\theta) = \E_{t, (\M x_0, \M x_1) \sim \pi} \left[ \left\| \M v_t^\theta(\M x_t, t) - (\M x_1 - \M x_0) \right\|_2^2 \right],
\end{equation}
The coupling $\pi(\mathbf{x}_0, \mathbf{x}_1)$ between marginals $p_{\mathbf{X}_0}$ and $p_{\mathbf{X}_1}$ fundamentally shapes the geometric properties of the resulting flow. While $\mathcal{L}_{\mathrm{CFM}}$ and $\mathcal{L}_{\mathrm{FM}}$ share equivalent gradients \citep{lipman2023flow}, the choice of $\pi$ determines the curvature of the probability paths. In practice, with $p_{\mathbf{X}_0}$ standardly defined as $\mathcal{N}(\mathbf{0}, \mathbf{I})$, an independent coupling $\pi = p_{\mathbf{X}_0} \otimes p_{\mathbf{X}_1}$ is straightforward to implement and yields a tractable conditional density $p_{\mathbf{X}_t | \mathbf{X}_1} = \mathcal{N}(t\mathbf{x}_1, (1-t)^2\mathbf{I})$. However, this approach often leads to intersecting trajectories and high-variance gradients during training. Conversely, an optimal transport (OT) coupling $\pi^\star$ minimizes the expected quadratic cost to produce straighter, non-overlapping paths along Wasserstein-2 geodesics \citep{peyre2019computational, pooladian2023multisample}. Although global OT is computationally prohibitive, mini-batch approximations facilitate faster convergence through local alignment \citep{tong2023conditional}, albeit at the cost of losing a simple closed-form characterization for $p_{\mathbf{X}_t | \mathbf{X}_1}$. Crucially, these straighter trajectories promote a more regular and near-linear mapping $\psi$, a property that we later exploit to ensure the stability and validity of our  proxy-gradient strategy.

\subsection{Source optimization through flows}
A powerful paradigm for solving inverse problems with generative flows is source optimization, as popularized by frameworks such as D-Flow \citep{ben2024d}. Given a pre-trained velocity field $\M v^\theta$ that defines a flow path $\psi_t$, the objective can be formulated in the latent source space:
\begin{equation}
    \min_{\mathbf{x}_0} \mathcal{L}(\psi_t(\mathbf{x}_0), \mathbf{y}) + \mathcal{R}(\mathbf{x}_0).
    \label{eq:source_opt}
\end{equation}
To solve this, gradient-based methods update the source point via $\mathbf{x}_0 \leftarrow \mathbf{x}_0 - \eta \nabla_{\mathbf{x}_0} \mathcal{L}$. By the chain rule, the gradient is computed as:
\begin{equation}
    \nabla_{\mathbf{x}_0} \mathcal{L} = \left( \frac{d\mathbf{x}_1}{d\mathbf{x}_0} \right)^\top \nabla_{\mathbf{x}_1} \mathcal{L}.
\end{equation}
Although this approach enables high-fidelity reconstruction, differentiating through the ODE solver requires an extensive chain of Jacobian matrices $\frac{\partial \mathbf{x}_{t+\delta t}}{\partial \mathbf{x}_t}$. This process is often plagued by numerical instability and prohibitive computational overhead, directly motivating the more efficient proxy-gradient strategy proposed in this work.

\subsection{Gaussian annulus theorem}\label{subsec:gaussian}
Instead of incorporating a regularization term into the optimization objective, we apply Gaussian spherical projection to maintain the prior consistency. This operation is grounded in the Gaussian annulus theorem \cite{vershynin2018high}, which characterizes the concentration of measure in high-dimensional spaces. For a $d$-dimensional Gaussian prior $z \sim \mathcal{N}(0, \mathbf{I})$, the probability mass concentrates on a thin shell of radius $\sqrt{d}$, known as the `soap bubble' effect. Optimization that neglects this property often drifts into low-probability regions, yielding unrealistic or blurry results \cite{menon2020pulse}. To maintain realism and sharpness, we constrain the latent vector $\M z$ to the high-density region of the prior by projecting it onto the hypersphere manifold $\mathcal{M}$:
\begin{equation}
    \mathcal{M} = \mathbb{S}^{d-1}(\sqrt{d}),
\end{equation}
where $\mathbb{S}^{n}(r) = \{ \mathbf{z} \in \mathbb{R}^{n+1} : \|\mathbf{z}\|_2 = r \}$ denotes an $n$-sphere of radius $r$ embedded in $(n+1)$-dimensional space. This projection ensures the latent variables remain consistent with the expected norm of the Gaussian prior \cite{bird2024}, preserving the perceptual quality of the reconstructed outputs.

\begin{figure}
    \centering
    \includegraphics[width=1\linewidth]{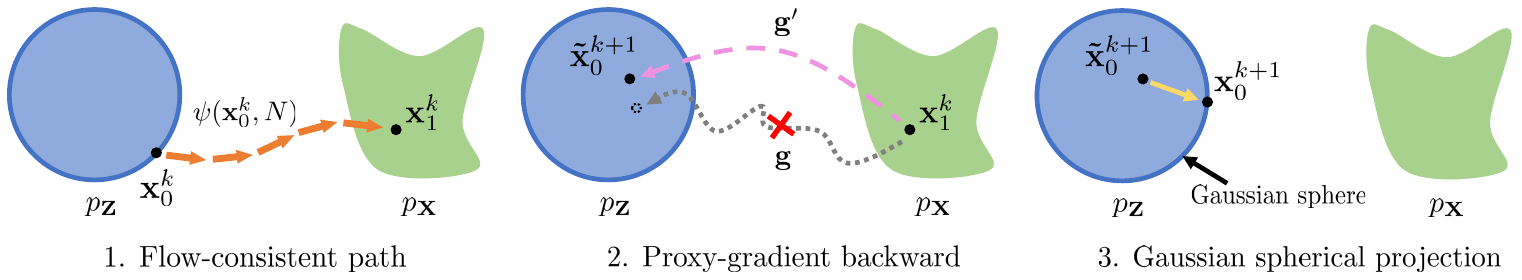}
    \caption{Overview of the P-Flow framework. The restoration follows a three-step cycle: (i) Flow-consistent path: generating $\mathbf{x}_1$ from latent source $\mathbf{x}_0$ via ODE integration; (ii) proxy gradient backward: updating $\mathbf{x}_0$ with a  gradient proxy to bypass unstable Jacobian chains; and (iii) Gaussian spherical projection: constraining the latent variable within the typical region of the Gaussian prior.}
    \label{fig:system}
\end{figure}

\section{Method}\label{sec:method}
\subsection{P-Flow: proxy-gradient source optimization}
We consider a pre-trained velocity field $\mathbf{v}_t^\theta$ that defines an ODE path transporting a source distribution $p_{\M X_0}$ to a target data distribution $p_{\M X_1}$. For a given degraded observation $\mathbf{y}$, our goal is to recover the sample by finding an optimal source point $\mathbf{x}_0$ in the latent space. Specifically, we iteratively optimize $\M x_0$ such that its corresponding generated sample $\M x_1$ minimizes a data-fidelity loss $\mathcal{L}(\mathbf{x}_1, \M y)$. As depicted in Figure~\ref{fig:system}, the optimization involves $K$ iterations, each comprising a three-step procedure:
\begin{enumerate}
    \item \textbf{Time-consistent flow}: 
    \begin{equation}
        \M x^k_1 = \psi(\M x_0^k, N).
    \end{equation}
    In this step, we obtain the generated sample $\M x_1^k$ by time-consistent flow $\psi_t$  where $N$ is ODE steps. This mapping represents a complete trajectory from the source distribution to the data manifold.
    \item \textbf{Gradient step on data-fidelity term}:
    \begin{equation}
        \mathbf{\tilde{x}}_0^{k+1} = \mathbf{x}_0^k - \eta \M g'.
    \end{equation}
    We update the source point by minimizing the data-fidelity term, where $\M g'=C\nabla_{\mathbf{x}_1} \mathcal{L}$ is a proxy gradient and $C$ is a proxy scalar. Notably, $C$ can be subsumed into $\eta$, thereby eliminating the need for an additional hyperparameter. This approach bypasses the numerical instability and computational overhead in differentiating through long-chain trajectories.
    \item \textbf{Gaussian spherical projection}: 
    \begin{equation}
        \mathbf{x}_0^{k+1} =  \frac{\sqrt{d}}{\|\mathbf{\tilde{x}}_0^{k+1}\|}\mathbf{\tilde{x}}_0^{k+1}.
    \end{equation}
    According to the Gaussian annulus theorem in Section \ref{subsec:gaussian}, the probability mass of a high-dimensional Gaussian distribution is concentrated on a hypersphere of radius $\sqrt{d}$. To prevent the optimized latent from drifting into low-probability regions and to ensure $\mathbf{x}_0$ remains a valid sample within the Gaussian prior, we project the updated vector back onto the hypersphere. Appendix~\ref{app:algorithm} provides the summarized algorithm.
\end{enumerate}

\subsection{Theoretical analysis}\label{subsec:theory}
\paragraph{Bayesian consistency of source optimization} 
Given the mapping $\mathbf{x}_1 = \psi(\mathbf{x}_0)$ defined by the flow, the log-prior $\log p_{\mathbf{X}_1}$ can be related to the latent source through: 
\begin{equation}
    \log p_{\mathbf{X}_1} = \log p_{\mathbf{X}_0} - \log | \det \nabla_{\mathbf{X}_0} \psi|. 
\end{equation}
Substituting into the MAP objective, the problem is reformulated in the latent space as minimizing:
\begin{equation}
    \mathcal{J}(\mathbf{x}_0) = -\log p(\mathbf{y}|\psi(\mathbf{x}_0)) - \log p_{\mathbf{X}_0} + \log | \det \nabla_{\mathbf{X}_0} \psi |. 
\end{equation}
Drawing upon the theoretical properties of FM, the OT displacement interpolants promote nearly straight probability paths with minimal kinetic energy \citep{lipman2023flow, liu2022flow}. As observed in recent flow-based inverse solvers \citep{pourya2026flower,ben2024d}, such straightness encourages a more regular and smoother mapping $\psi$ and exhibits a relatively stable Jacobian determinant. Consequently, the term $\log | \det \nabla_{\mathbf{x}_0} \psi |$ varies minimally across the typical regions of the latent space and can be treated as an approximately constant offset. By neglecting this constant term, the objective simplifies to a latent-space optimization problem:
\begin{equation}
    \hat{\mathbf{x}}_0 = \arg\min_{\mathbf{x}_0} \left\{ \mathcal{L}(\psi(\mathbf{x}_0), \mathbf{y}) + \frac{\lambda}{2} \|\mathbf{x}_0\|^2 \right\},
\end{equation}
where the Gaussian prior $p_{\mathbf{X}_0}$ provides an explicit $\ell_2$ regularization. However, the tractability of this reformulated optimization landscape critically depends on the regularity of the mapping $\psi$. Specifically, for gradient-based methods to converge reliably, the mapping must guarantee that small perturbations in the latent source $\mathbf{x}_0$ do not lead to erratic explosions in the generated sample $\mathbf{x}_1$. This stability is formally characterized by the Lipschitz continuity of the flow. To establish this necessary condition for stable source optimization, we present the following proposition:
\begin{proposition}\label{prop:Lipschitz continuity}
Consider the flow path $\psi$ defined by the ODE $\frac{d}{dt}\mathbf{x}_t = \M v_t(\mathbf{x}_t)$ for $t \in [0, 1]$ with the initial condition $\mathbf{x}_{t=0} = \mathbf{x}_0$. If the velocity field $\M v_t$ is Lipschitz continuous with respect to $\mathbf{x}$, then the mapping $\psi$ is also Lipschitz continuous.
\end{proposition}
The proof can be found in Appendix \ref{app:Proof of proposition 1}. This consistency ensures that the source optimization is a theoretically grounded process. By transforming the intractable pixel-space prior into a well-defined Gaussian prior, we provide a rigorous starting point for the subsequent proxy-gradient strategy.

\paragraph{Numerical instability of long-chain differentiation.}
The fundamental challenge in source optimization lies in the numerical instability of long-chain differentiation through ODE trajectories. We denote $\M M_N$ as the cumulative Jacobian of the ODE trajectory:
\begin{equation}
    \mathbf{M}_N = \frac{d\mathbf{x}_1}{d\mathbf{x}_0} = \prod_{i=0}^{N-1} \frac{\partial \mathbf{x}_{t_{i+1}}}{\partial \mathbf{x}_{t_i}} = \prod_{i=0}^{N-1} \left( \mathbf{I} + \delta t \mathbf{J}_i \right).
    \label{eq:full_jacobian}
\end{equation}
In deep neural networks, the cumulative product of Jacobians induces a phenomenon known as spectral accumulation \citep{ben2024d, pascanu2013difficulty}, where the concentration of singular values leads to severe ill-conditioning of the gradient path \cite{pokle2023training}. This effect is rooted in the inherent spectral anisotropy of neural network Jacobians: due to the non-linear activation functions and the manifold-learning objective, the field exhibits non-uniform spectral scaling across the spatial domain \cite{saxe2013exact}. Instead, the local Jacobian $\mathbf{J}_i$ tends to have a highly non-uniform singular value distribution, which is consistently observed in deep architectures and leads to the degradation of gradient signals \cite{ pascanu2013difficulty,balduzzi2017shattered}. This persistent deviation from isotropy at each infinitesimal step compounds over the ODE trajectory, as formalized below:
\begin{proposition} \label{prop:ill_conditioning}
Let $\kappa(\mathbf{M}_N)$ denote the condition number of the cumulative Jacobian. Given the spectral anisotropy of velocity fields where each local step $\mathbf{A}_i = \mathbf{I} + \delta t \mathbf{J}_i$ possesses a condition number $\kappa(\mathbf{A}_i) > 1$. Under persistent spectral anisotropy and aligned principal singular directions, the cumulative condition number $\kappa(\mathbf{M}_N)$ can grow rapidly with the ODE step $N$.

\end{proposition}
The proof can be found in Appendix \ref{app:Proof of proposition 2}. Consequently, the gradient $\M g=\nabla_{\M x_0}\mathcal{L}$ collapses onto a few dominant singular directions, obscuring essential signals in other dimensions with numerical noise. This creates an irregular optimization landscape that is difficult for first-order solvers to navigate, demonstrating that the instability is a fundamental structural problem inherent in differentiating through long ODE trajectories. We further provide perturbation analysis in Appendix~\ref{app:Numerical perturbation}.

\paragraph{Convergence analysis}
By Proposition 1, the flow path $\psi$ inherits Lipschitz continuity from the velocity field $\mathbf{v}_t$. To circumvent the numerical instability of the true gradient $\mathbf{g} = \mathbf{M}_N^\top \nabla_{\mathbf{x}_1}\mathcal{L}$, we employ a proxy gradient $\mathbf{g}' = C\nabla_{\mathbf{x}_1}\mathcal{L}$, where $C > 0$ can be absorbed into the learning rate $\eta$.
\par
To characterize when the proxy gradient is valid, we first give a sufficient condition. If the symmetric directional deformation induced by the flow does not reverse the loss-gradient direction, then the proxy gradient is a descent direction. The stronger condition $\|\mathbf{M}_N - \mathbf{I}\| < 1$ is one simple sufficient case, but P-Flow only requires a weaker positive directional alignment condition. 
\begin{proposition}
\label{prop:convergence}
Let $\mathbf{u} = \nabla_{\mathbf{x}_1}\mathcal{L}(\mathbf{x}_1, \mathbf{y})$ be the loss gradient. Assume the cumulative Jacobian $\mathbf{M}_N$ satisfies the directional positive alignment condition $\mathbf{u}^\top \mathbf{M}_N \mathbf{u} \ge \alpha \|\mathbf{u}\|^2$ for some constant $\alpha > 0$. Assuming the composite objective $J(\mathbf{x}_0) = \mathcal{L}(\psi(\mathbf{x}_0))$ is $L_J$-smooth, the iterative update $\mathbf{x}_0^{k+1} = \mathbf{x}_0^k - \eta \mathbf{g}'$ converges to a stationary point for a sufficiently small step size $\eta$.
\end{proposition}
The proof can be found in Appendix A.4. The accumulated Jacobian $\mathbf{M}_N$ may be ill-conditioned, which makes exact backpropagation numerically unstable. However, P-Flow does not require $\mathbf{M}_N$ to be well-conditioned. It only requires the proxy direction to maintain a positive directional alignment with the exact gradient along the optimization trajectory. This weaker condition is theoretically sufficient for descent and is empirically validated by the cosine-alignment measurements in Figure 5, where $\cos\angle(\mathbf{g}, \mathbf{g}') > 0$ directly verifies the empirical version of $\alpha > 0$.
\begin{remark}
Beyond providing a descent direction, the projection in Step 3 makes $\M x_0$ remain on the high-probability manifold $\mathcal{M}$. Grounded in the Gaussian annulus theorem, this constraint acts as a regularization to prevent numerical drift into low-prior regions, which is essential for stabilizing the optimization.
\end{remark}

\subsection{Complexity analysis}\label{sec:complexity}
P-Flow significantly reduces computational overhead compared to full-chain differentiation methods like D-Flow \citep{ben2024d}. Let $M^{\theta}$ denote the memory for a single forward pass, and $T_{f}, T_{b}, T_{\M g}$ represent the time costs for a forward pass, a backward pass, and the data-fidelity gradient $\nabla_{\mathbf{x}_1} \mathcal{L}$, respectively. Traditional methods require caching intermediate activations for all $N$ steps, leading to $\mathcal{O}(N \cdot M^{\theta})$ memory. In contrast, P-Flow decouples optimization from ODE dynamics, requiring only $\mathcal{O}(M^{\theta})$ memory, which is independent of $N$. Regarding time complexity, while standard methods incur $\mathcal{O}(N \cdot (T_{f} + T_{b}))$ latency per iteration due to backpropagation through the unrolled trajectory, P-Flow reduces this to $\mathcal{O}(N \cdot T_{f} + T_{\M g})$. Since $T_{\M g} \ll T_{b}$ for common linear operators, P-Flow achieves a substantial reduction in inference latency and significantly higher throughput.

\section{Related works}\label{sec:related works}
Recent advances in inverse problems leverage pretrained diffusion and flow models as generative priors, achieving zero-shot restoration by integrating these priors with iterative guidance to maintain data consistency. We categorize these methods into three methodological paradigms:
\paragraph{Conditioned sampling and posterior guidance} 
This stream focuses on steering the generative trajectory toward the observation $\mathbf{y}$. DPS \citep{chung2023diffusion} pioneered gradient-based guidance via the measurement residual $\nabla_{\mathbf{x}_t} \|\mathbf{y} - \mathbf{H}\hat{\mathbf{x}}_0\|^2$. To improve spectral handling, DDRM \citep{kawar2023denoising} utilizes singular value decomposition in the spectral domain, while $\Pi$GDM \citep{song2023pseudoinverse} extends this to non-linear degradations using pseudoinverse guidance. More recently, DAPS \citep{zhang2025improving} mitigates global error accumulation by decoupling consecutive samples through recursive marginal sampling from $p(\mathbf{x}_t | \mathbf{y})$.
\paragraph{Variational integration and PnP frameworks} 
These methods unify generative models with classical optimization frameworks like PnP. DiffPIR \citep{zhu2023denoising} alternates between data-fidelity proximal steps and diffusion denoising, while PnP-GS \citep{hurault2021gradient} ensures theoretical convergence by parameterizing denoisers as gradient descent steps. Within the FM framework, PnP-Flow \citep{martin2025pnpflow} employs a time-dependent denoiser for trajectory reprojection. Similarly, Flower \citep{pourya2026flower} unifies these views by refining flow-consistent destinations via proximal operators $\mu_t = \text{prox}_{\nu_t^2 F_y} (\hat{\mathbf{x}}_1)$ before re-projecting them onto the flow path.
\paragraph{Source optimization and numerical preconditioning} 
A third category seeks to reduce the computational cost of solving the inverse problem. D-Flow \citep{ben2024d} formulates the task as a source point optimization $\min_{\mathbf{x}_0} \mathcal{L}(\mathbf{x}_1)$, though it requires expensive differentiation through unrolled ODEs. In diffusion models, BIRD \citep{bird2024} similarly casts the task as an initial noise optimization, mitigating the prohibitive unrolling costs by adopting large time steps to accelerate sampling. To enhance efficiency, ICTM \citep{zhang2024flow} approximates the MAP objective using local objectives $\sum \gamma_i \hat{J}_i$ to avoid full backpropagation. Numerical stability is further addressed by DDPG \citep{garber2024image} through an iterative preconditioning matrix $\mathbf{W}_t$ that balances consistency and noise robustness. Finally, DDS \citep{chung2023decomposed} accelerates high-dimensional problems by replacing manifold-constrained gradients with efficient multi-step conjugate gradient updates in the Krylov subspace.

\section{Numerical results}\label{sec:numerical results}

\subsection{Experimental setup}\label{sec:experimental setup}

\paragraph{Datasets}
We evaluate the proposed method on two datasets: FFHQ \cite{karras2019style}, and AFHQ-Cat (the cat subset of AFHQ \cite{choi2020stargan}). All images are processed at $256 \times 256$ resolution and normalized to the range $[-1, 1]$. Since FFHQ does not provide an official validation set, we randomly select a subset of 128 images for evaluation. For AFHQ-cat, we also randomly select 128 images from the test set for validation.

\paragraph{Models}
We employ a U-Net backbone consistent with recent FM models \citep{ho2020denoising, dhariwal2021diffusion}. The architecture follows the design in \cite{ronneberger2015u}, consisting of a series of residual blocks \cite{he2016deep} and self-attention layers \cite{vaswani2017attention} at lower resolutions. The network is conditioned on the time step $t$ via group normalization layers. For the velocity field $\mathbf{v}^\theta$ in our FM framework, the U-Net takes the noisy intermediate state $\mathbf{x}_t$ and the time $t$ as inputs to predict the corresponding vector field. Our models are trained using the CFM objective \cite{lipman2023flow}. We use the Adam optimizer \cite{kingma2014adam} with a constant learning rate of $10^{-4}$ and no weight decay. For both FFHQ and AFHQ-cat datasets, the model is trained for 200 epochs with a batch size of 16. For baseline evaluations, we train the denoising network for PnP-GS \citep{hurault2021gradient} employing the same U-Net architecture to ensure consistency. For DiffPIR \citep{zhu2023denoising}, we utilize the pre-trained model from \citep{choi2021ilvr} on the FFHQ dataset \cite{karras2019style}, implemented in the DeepInv library \cite{tachella2025deepinv}. 

\begin{table}[t]
    \caption{Results on 128 test images of the FFHQ dataset.}
    \label{tab:benchmark_results_ffhq}
    \centering
    \setlength{\tabcolsep}{2pt}
    \scriptsize
    \resizebox{1\textwidth}{!}{
        \begin{tabular}{lccccccccccccccc}
            \toprule
            \multirow{3}{*}{Method}
              & \multicolumn{3}{c}{Denoising}
              & \multicolumn{3}{c}{Deblurring}
              & \multicolumn{3}{c}{Super-resolution}
              & \multicolumn{3}{c}{Random inpainting}
              & \multicolumn{3}{c}{Box inpainting} \\
            \cmidrule(lr){2-4}\cmidrule(lr){5-7}\cmidrule(lr){8-10}\cmidrule(lr){11-13}\cmidrule(lr){14-16}
            & PSNR & SSIM & LPIPS & PSNR & SSIM & LPIPS & PSNR & SSIM & LPIPS & PSNR & SSIM & LPIPS & PSNR & SSIM & LPIPS \\
            \midrule
            Degraded      & 20.45 & 0.303 & 0.484 & 23.58 & 0.551 & 0.422 & 11.14 & 0.215 & 0.868 & 12.75 & 0.215 & 1.078 & 22.13 & 0.742 & 0.200 \\
            PnP-GS        & 31.00 & 0.819 & 0.088 & 26.61 & 0.784 & 0.259 & 19.61 & 0.534 & 0.432 & 23.54 & 0.753 & 0.214 & - & - & - \\
            DiffPIR       & 31.90 & 0.876 & 0.091 & \textbf{29.41} & \underline{0.824} & 0.165 & 25.14 & 0.726 & 0.194 & 32.73 & 0.911 & \underline{0.032} & - & - & - \\
            OT-ODE        & 30.78 & 0.850 & \underline{0.079} & 27.45 & 0.772 & \textbf{0.112} & 26.07 & \underline{0.772} & \textbf{0.095} & 29.52 & 0.867 & 0.078 & 27.94 & 0.897 & 0.069 \\
            Flow-Priors   & 31.45 & 0.873 & 0.097 & 26.89 & 0.756 & 0.178 & 20.93 & 0.482 & 0.361 & 32.48 & 0.906 & \underline{0.032} & 29.56 & 0.834 & 0.082 \\
            PnP-Flow      & 32.29 & 0.892 & 0.117 & 28.01 & 0.806 & 0.215 & 24.58 & 0.738 & 0.179 & \textbf{33.84} & \textbf{0.938} & 0.038 & \underline{30.03} & \textbf{0.934} & \underline{0.038} \\
            Flower        & \textbf{32.54} & \textbf{0.900} & 0.080 & \underline{29.18} & \textbf{0.827} & 0.178 & 24.27 & 0.748 & 0.198 & 33.07 & \underline{0.929} & \textbf{0.030} & 29.55 & 0.927 & \textbf{0.034} \\
            D-Flow        & 26.22 & 0.558 & 0.178 & 28.09 & 0.793 & 0.175 & 23.99 & 0.615 & 0.194 & 32.78 & 0.916 & 0.038 & 29.32 & 0.865 & 0.053 \\
            P-Flow (ours) & \underline{32.45} & \underline{0.894} & \textbf{0.078} & 28.07 & 0.788 & \underline{0.143} & \textbf{27.64} & \textbf{0.819} & \underline{0.132} & \underline{33.29} & 0.923 & 0.050 & \textbf{30.30} & \underline{0.930} & 0.055 \\
            \bottomrule
        \end{tabular}
    }
\end{table}

\begin{table}[t]
    \caption{Results on 128 test images of the AFHQ-Cat dataset.}
    \label{tab:benchmark_results_cats}
    \centering
    \setlength{\tabcolsep}{2pt} 
    \scriptsize 
    \resizebox{1\textwidth}{!}{
        \begin{tabular}{lccccccccccccccc}
            \toprule
            \multirow{3}{*}{Method}
              & \multicolumn{3}{c}{Denoising}
              & \multicolumn{3}{c}{Deblurring}
              & \multicolumn{3}{c}{Super-resolution}
              & \multicolumn{3}{c}{Random inpainting}
              & \multicolumn{3}{c}{Box inpainting} \\
            \cmidrule(lr){2-4}\cmidrule(lr){5-7}\cmidrule(lr){8-10}\cmidrule(lr){11-13}\cmidrule(lr){14-16}
            & PSNR & SSIM & LPIPS & PSNR & SSIM & LPIPS & PSNR & SSIM & LPIPS & PSNR & SSIM & LPIPS & PSNR & SSIM & LPIPS \\
            \midrule
            Degraded      & 20.00 & 0.314 & 0.509 & 23.94 & 0.517 & 0.444 & 11.70 & 0.208 & 0.873 & 13.36 & 0.223 & 1.081 & 21.80 & 0.740 & 0.198 \\
            PnP-GS        & \textbf{32.58} & \textbf{0.893} & \underline{0.070} & 27.88 & 0.753 & 0.350 & 24.10 & 0.642 & 0.362 & 29.22 & 0.829 & 0.120 & - & - & - \\
            DiffPIR       & 30.91 & 0.837 & 0.189 & 29.21 & 0.781 & 0.273 & 24.17 & 0.635 & 0.392 & 32.14 & 0.884 & 0.058 & - & - & - \\
            OT-ODE        & 30.52 & 0.832 & 0.100 & 27.67 & 0.730 & \textbf{0.123} & 26.56 & 0.734 & \textbf{0.107} & 29.93 & 0.849 & 0.085 & 24.67 & 0.876 & 0.092 \\
            Flow-Priors   & 31.64 & 0.877 & \textbf{0.068} & 27.60 & 0.741 & 0.188 & 25.41 & 0.584 & 0.281 & 33.16 & 0.900 & 0.065 & 26.43 & 0.811 & 0.139 \\
            PnP-Flow      & 31.76 & 0.867 & 0.172 & \textbf{29.83} & \underline{0.800} & 0.299 & \underline{27.49} & \textbf{0.780} & \underline{0.154} & \textbf{33.90} & \textbf{0.925} & 0.053 & \underline{27.43} & \textbf{0.919} & \underline{0.070} \\
            Flower        & 32.03 & \underline{0.881} & 0.112 & \underline{29.64} & \textbf{0.802} & 0.230 & 25.21 & 0.713 & 0.249 & \underline{33.19} & \underline{0.915} & \textbf{0.042} & 26.68 & \underline{0.915} & \textbf{0.064} \\
            D-Flow        & 26.44 & 0.579 & 0.176 & 28.62 & 0.758 & \underline{0.160} & 25.10 & 0.616 & 0.187 & 32.64 & 0.898 & \underline{0.044} & 27.03 & 0.842 & 0.080 \\
            P-Flow (ours) & \underline{32.09} & 0.880 & 0.106 & 28.01 & 0.741 & 0.165 & \textbf{28.16} & \underline{0.777} & 0.157 & 32.73 & 0.900 & 0.060 & \textbf{27.56} & 0.908 & 0.081 \\
            \bottomrule
        \end{tabular}}
\end{table}

\paragraph{Settings for the tasks}
We evaluate the methods using 128 test images from both FFHQ and AFHQ-Cat datasets across five restoration tasks. Given that both datasets share a consistent resolution of $256\times 256$, we apply identical degradation configurations: (i) denoising with Gaussian noise ($\sigma=0.2$); (ii) deblurring using a $61\times 61$ Gaussian kernel with $\sigma_b=3.0$; (iii) $4\times$ super-resolution via bicubic downsampling; (iv) box-inpainting with a centered $80\times 80$ mask; and (v) random pixel inpainting with $70\%$ masked pixels. For the deblurring, super-resolution, and box-inpainting tasks, we further add Gaussian noise with $\sigma=0.05$. For random inpainting, the noise level is $\sigma=0.01$.
\begin{figure}[t!]
    \centering
    \includegraphics[width=0.95\linewidth]{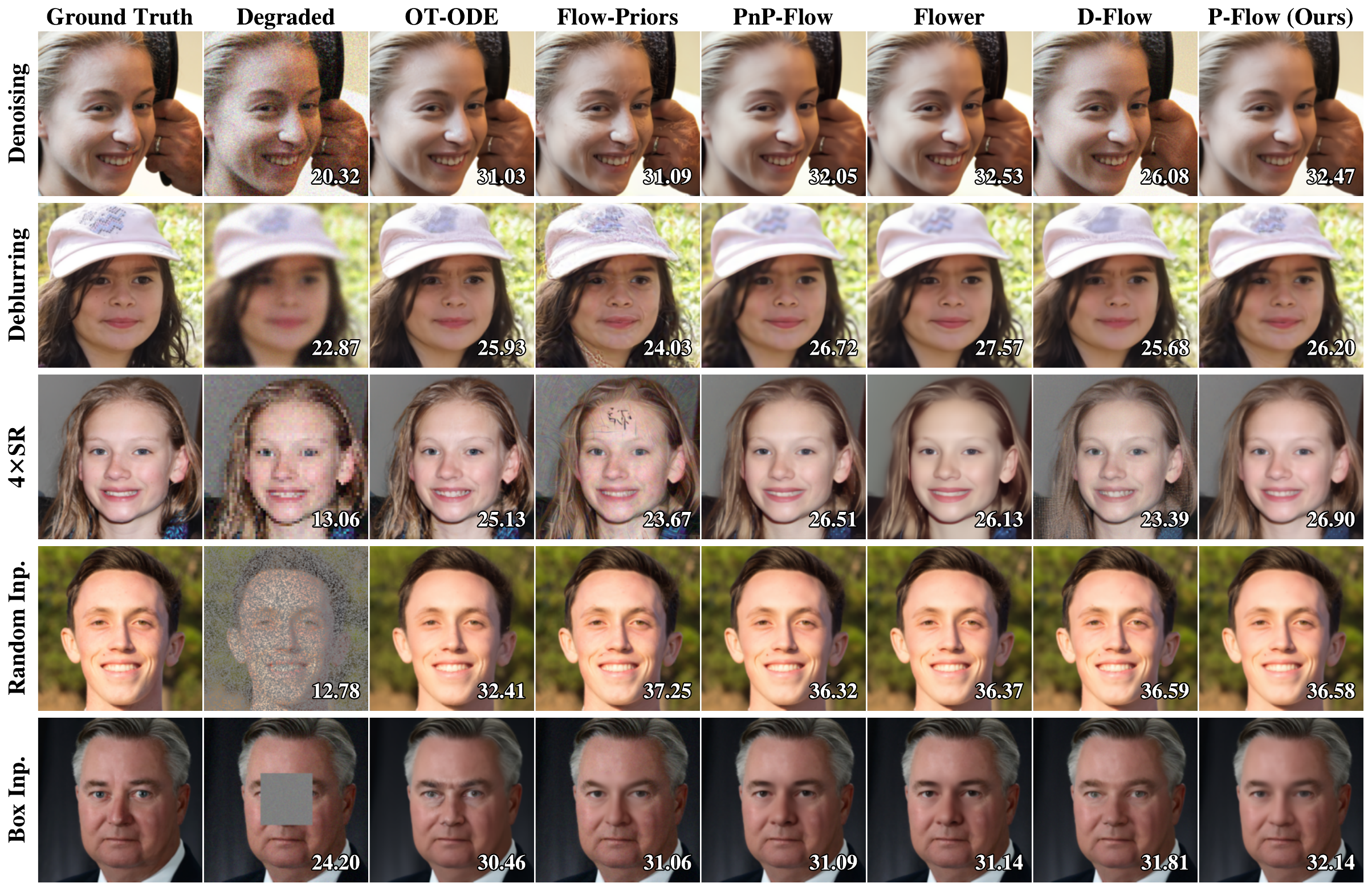}
    \caption{Qualitative comparison of image restoration results on the FFHQ dataset. The PSNR metrics are displayed in the bottom-right corner of each restored image.}
    \label{fig:Qualitative comparison on ffhq}
\end{figure}

\begin{figure}[t!]
    \centering
    \includegraphics[width=0.95\linewidth]{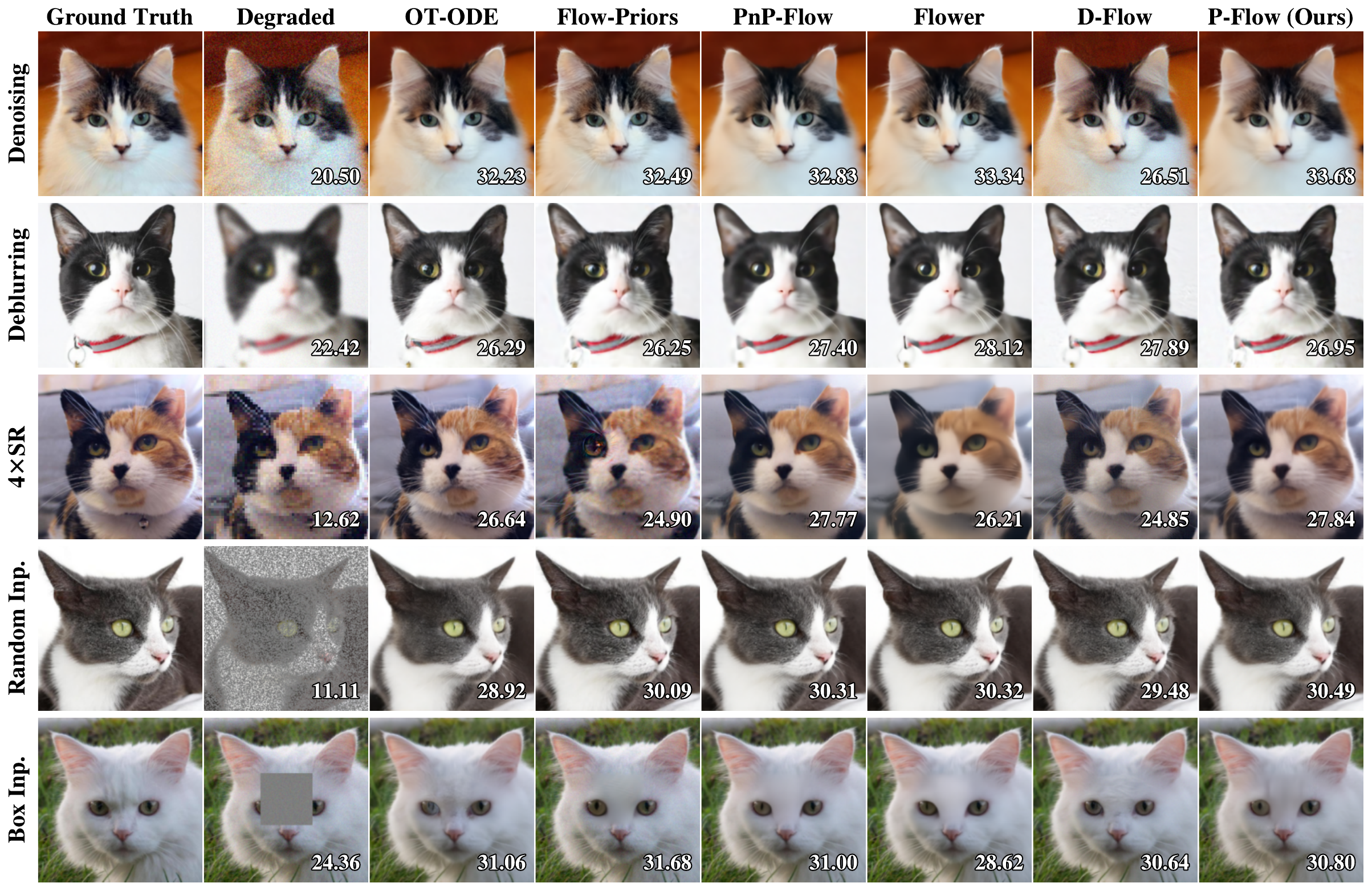}
    \caption{Qualitative comparison of image restoration results on the AFHQ-Cat dataset. The PSNR metrics are displayed in the bottom-right corner of each restored image.}
    \label{fig:Qualitative comparison on afhq_cat}
\end{figure}
\paragraph{Baselines}
To evaluate the performance of our method, we compare it against a wide range of baselines, including PnP-GS \cite{hurault2021gradient}, DiffPIR \cite{zhu2023denoising}, OT-ODE \cite{pokle2023training}, D-Flow \cite{ben2024d}, Flow-Priors \cite{zhang2024flow}, PnP-Flow \cite{martin2025pnpflow}, and Flower \cite{pourya2026flower}. To ensure fair comparisons, we conduct an extensive grid search for all methods to identify the optimal configurations (provided in Appendix~\ref{app:Implemention details}). For PnP-Flow and Flower, we restrict them to a single evaluation per image (setting the number of averagings $N_{Avg} = 1$) to ensure a fair comparison. For the Flower method specifically, we set the uncertainty hyperparameter to $\gamma = 0$, following the observation in \citep{pourya2026flower} that ignoring destination uncertainty leads to better reconstruction quality. All benchmarks are conducted on an NVIDIA A100 SXM 40GB GPU.

\subsection{Main results}
Table~\ref{tab:benchmark_results_ffhq} and Table~\ref{tab:benchmark_results_cats} benchmark the performance of P-Flow against other baselines. The PSNR, SSIM, and LPIPS metrics are averaged over 128 test images. Note that `-' in the table signifies that a particular method is inapplicable with the corresponding task; for instance, PnP-GS and DiffPIR are omitted from box inpainting evaluations due to their specific architectural constraints. Figure~\ref{fig:Qualitative comparison on ffhq} and Figure~\ref{fig:Qualitative comparison on afhq_cat} show qualitative comparisons of the restored images with their individual PSNR values for reference. Evaluation results indicate that our method maintains robust performance across all restoration tasks and achieves state-of-the-art results in several categories. 
\par
Compared to all baselines, P-Flow exhibits distinct characteristics depending on the specific task. For super-resolution and box inpainting, our method achieves consistently state-of-the-art performance. For the deblurring task, we acknowledge that P-Flow yields a relative shortcoming in reconstruction fidelity. However, it achieves competitive perceptual quality and more realistic texture recovery that better aligns with human visual perception. In standard denoising and random inpainting tasks, the performance of P-Flow is comparable to or slightly exceeds that of PnP-Flow and Flower. While our method may not dominate every quantitative metric under these standard settings, its strength dominates when conditions become severely ill-posed. As detailed in Appendix~\ref{app:Severely ill-posed conditions} and Appendix~\ref{app:With high measurement noise}, under extreme degraded conditions, baseline methods suffer from severe structural distortion, catastrophic color shifts, or total perceptual collapse. By contrast, P-Flow maintains remarkable stability and structural coherence even in these extreme regimes.
\par
Overall, P-Flow effectively balances reconstruction fidelity and perceptual realism. This is because the final generation operates as a pure ODE path decoupled from data-fidelity constraints, preventing intrusive interventions from disrupting the natural data manifold. Extensive evaluations are provided in the Appendix, including further visualizations (Appendix~\ref{app:Additional visual results}), ablations on the Gaussian projection (Appendix~\ref{app:Gaussian spherical projection}), hyperparameter sensitivities for $N$ and $K$ and $\eta$ (Appendix~\ref{app:Impact of ODE steps}, \ref{Impact of interation numbers}, \ref{app:learning rate}), and computational efficiency benchmarks (Appendix~\ref{app:Computation time and memory}).

\section{Conclusion}\label{sec:conclusion}
In this work, we presented P-Flow, an efficient and stable framework for linear inverse problems based on flow matching. By introducing a proxy-gradient strategy, P-Flow effectively bypasses the numerical instabilities and prohibitive memory overhead associated with differentiating through unrolled ODE trajectories. To maintain latent source consistency, we incorporated a Gaussian spherical projection grounded in the concentration of measure, keeping the optimization within the typical region of the prior. Theoretical analysis and extensive experiments across diverse tasks demonstrate that P-Flow achieves competitive restoration quality while significantly enhancing optimization stability and computational efficiency.

\section{Limitations}\label{sec:Limitation}
Despite the significant efficiency gains over full-chain differentiation, P-Flow still faces a latency bottleneck, as each optimization iteration necessitates a complete forward integration of the ODE trajectory. Furthermore, our current analysis primarily focuses on linear inverse problems; the effectiveness and theoretical stability of the proxy-gradient strategy in the context of complex nonlinear forward operators remain to be more extensively explored in future research.

{
\small
\bibliographystyle{unsrt}
\bibliography{reference}
}


\appendix

\section{Proof}\label{app:proof}
\subsection{Proof of proposition 1}\label{app:Proof of proposition 1}
We begin by assuming the Lipschitz continuity of the learned velocity field $\M v_t$, which is a standard property of neural networks with bounded weights and smooth activation functions.
\par
We assume there exists a finite constant $L_{\M v} > 0$ such that for any $\mathbf{x}, \mathbf{x}' \in \mathbb{R}^d$ and any $t \in [0, 1]$, the velocity field satisfies:
\begin{equation}
    \|\M v_t(\mathbf{x}) - \M v_t(\mathbf{x}')\| \le L_{\M v} \|\mathbf{x} - \mathbf{x}'\|.
\end{equation}

Now, consider two distinct initial source points $\mathbf{x}_0$ and $\mathbf{x}'_0$ in the latent space. These points induce two trajectories, $\mathbf{x}_t$ and $\mathbf{x}'_t$, respectively, governed by the same velocity field $\M v_t$. The state at any time $t \in [0, 1]$ can be expressed in the integral form of the ODE:
\begin{equation}
    \mathbf{x}_t = \mathbf{x}_0 + \int_0^t \M v_s(\mathbf{x}_s) ds, \quad \mathbf{x}'_t = \mathbf{x}'_0 + \int_0^t \M v_s(\mathbf{x}'_s) ds.
\end{equation}
The difference between the two trajectories at time $t$ is:
\begin{equation}
\begin{aligned}
    \|\mathbf{x}_t - \mathbf{x}'_t\| &= \left\| (\mathbf{x}_0 - \mathbf{x}'_0) + \int_0^t \left( \mathbf{v}_s(\mathbf{x}_s) - \mathbf{v}_s(\mathbf{x}'_s) \right) ds \right\| && (\text{Integral form of ODE}) \\
    &\le \|\mathbf{x}_0 - \mathbf{x}'_0\| + \int_0^t \left\| \mathbf{v}_s(\mathbf{x}_s) - \mathbf{v}_s(\mathbf{x}'_s) \right\| ds && (\text{Triangle inequality}) \\
    &\le \|\mathbf{x}_0 - \mathbf{x}'_0\| + L_{\mathbf{v}} \int_0^t \|\mathbf{x}_s - \mathbf{x}'_s\| ds && (\text{Lipschitz property of } \mathbf{v}_t) \\
    &\le \|\mathbf{x}_0 - \mathbf{x}'_0\| \exp(L_{\mathbf{v}} t) && (\text{Grönwall's inequality})
\end{aligned}
\label{eq:gronwall_derivation}
\end{equation}
At the end of the trajectory ($t=1$), the generated samples are $\mathbf{x}_1 = \psi(\mathbf{x}_0)$ and $\mathbf{x}'_1 = \psi(\mathbf{x}'_0)$. Substituting $t=1$ into the above inequality yields:
\begin{equation}
    \|\psi(\mathbf{x}_0) - \psi(\mathbf{x}'_0)\| \le \exp(L_{\M v}) \|\mathbf{x}_0 - \mathbf{x}'_0\|.
\end{equation}
Let $L_\psi = \exp(L_{\M v})$. Since $L_{\M v}$ is a finite constant determined by the neural network's spectral norm, $L_\psi$ is also a finite constant. Thus, the mapping $\psi$ is $L_\psi$-Lipschitz continuous.

\subsection{Proof of proposition 2}\label{app:Proof of proposition 2}
Let $\mathbf{M}_N = \prod_{i=0}^{N-1} \mathbf{A}_i$, where $\mathbf{A}_i = \mathbf{I} + \delta t \mathbf{J}_i \in \mathbb{R}^{d \times d}$ represents the transition matrix at step $i$. Let $\sigma_1(\mathbf{A}) \ge \sigma_2(\mathbf{A}) \ge \dots \ge \sigma_d(\mathbf{A}) > 0$ denote the singular values of a matrix $\mathbf{A}$. The condition number is defined as $\kappa(\mathbf{A}) = \sigma_1(\mathbf{A}) / \sigma_d(\mathbf{A})$.

By the multiplicative properties of singular values, for any two matrices $\mathbf{A}$ and $\mathbf{B}$, we have the following lower bound for the maximum singular value and upper bound for the minimum singular value:
\begin{equation}
    \sigma_1(\mathbf{A}\mathbf{B}) \ge \sigma_1(\mathbf{A})\sigma_d(\mathbf{B}), \quad \sigma_d(\mathbf{A}\mathbf{B}) \le \sigma_d(\mathbf{A})\sigma_1(\mathbf{B}).
\end{equation}
Applying these properties recursively to the product $\mathbf{M}_N$, we obtain:
\begin{equation}
    \sigma_1(\mathbf{M}_N) \ge \sigma_1(\mathbf{A}_{N-1}) \prod_{i=0}^{N-2} \sigma_d(\mathbf{A}_i), \quad \sigma_d(\mathbf{M}_N) \le \sigma_d(\mathbf{A}_{N-1}) \prod_{i=0}^{N-2} \sigma_1(\mathbf{A}_i).
\end{equation}
However, a more direct lower bound for the condition number growth can be derived using the inequality $\kappa(\mathbf{A}\mathbf{B}) \ge \frac{\kappa(\mathbf{A})}{\kappa(\mathbf{B}^{-1})}$ (under certain alignment conditions) or more generally via the spectral radius. In the context of deep neural networks, let us assume that each local block $\mathbf{A}_i$ exhibits a consistent degree of anisotropy, such that $\kappa(\mathbf{A}_i) \ge 1 + \epsilon$ for some $\epsilon > 0$. According to the spectral growth theorem for matrix products, the cumulative condition number satisfies:
\begin{equation}
    \kappa(\mathbf{M}_N) \ge \prod_{i=0}^{N-1} \gamma_i,
\end{equation}
where $\gamma_i$ is related to the alignment of the principal singular vectors of $\{\mathbf{A}_i\}$. In the worst-case scenario (aligned singular vectors), $\kappa(\mathbf{M}_N) = \prod_{i=0}^{N-1} \kappa(\mathbf{A}_i) \ge (1+\epsilon)^N$. 
\par

\begin{figure}[h]
    \centering
    \includegraphics[width=0.95\linewidth]{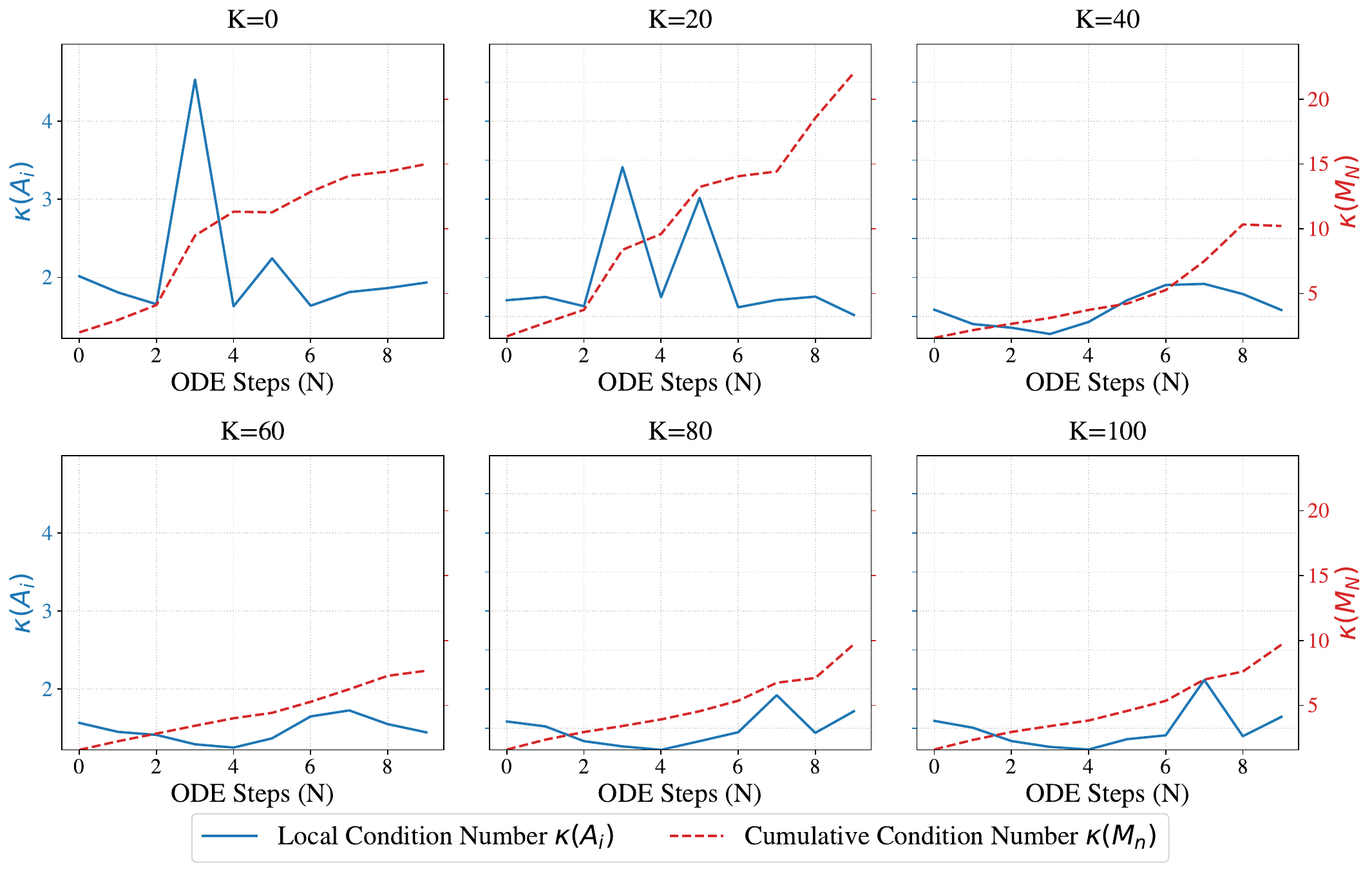}
    \caption{Evolution of local and cumulative condition numbers during ODE integration. Each panel illustrates the trends of the local condition number $\kappa(\mathbf{A}_i)$ (blue) and the cumulative condition number $\kappa(\mathbf{M}_N)$ (red) across ODE steps ($N$) at various optimization iterations.}
    \label{fig:jacobian_grid}
\end{figure}

In cases where singular vectors exhibit persistent alignment, $\kappa(\mathbf{M}_N)$ can exhibit rapid growth. This exponential explosion implies that for a large $N$, the gradient $\nabla_{\mathbf{x}_0} \mathcal{L} = \mathbf{M}_N^\top \nabla_{\mathbf{x}_1} \mathcal{L}$ becomes numerically singular, where any component of the gradient not aligned with the dominant singular vector of $\mathbf{M}_N$ is exponentially suppressed. We provide an empirical validation in Figure~\ref{fig:jacobian_grid}, where we visualize the exponential evolution of the cumulative condition numbers.

\subsection{Numerical perturbation}\label{app:Numerical perturbation}
The instability is further compounded by numerical noise. Consider a perturbed gradient $\tilde{\mathbf{g}}_1 = \nabla_{\mathbf{x}_1} \mathcal{L} + \boldsymbol{\xi}$, where $\boldsymbol{\xi}$ represents a small numerical error (e.g., floating-point noise). The error in the back-propagated gradient $\tilde{\mathbf{g}}_0$ is:
\begin{equation}
    \Delta \mathbf{g}_0 = \mathbf{M}_N^\top \boldsymbol{\xi}.
\end{equation}
The relative error is bounded as:
\begin{equation}
    \frac{\|\Delta \mathbf{g}_0\|}{\|\mathbf{g}_0\|} \le \kappa(\mathbf{M}_N) \frac{\|\boldsymbol{\xi}\|}{\|\nabla_{\mathbf{x}_1} \mathcal{L}\|}.
\end{equation}
Since $\kappa(\mathbf{M}_N)$ grows as $\mathcal{O}((1+\epsilon)^N)$, even a machine-epsilon level perturbation $\boldsymbol{\xi}$ can result in a relative error of order $1$ or greater for large $N$. This effectively `shatters' the gradient, rendering the descent direction stochastic and uninformative.

\subsection{Proof of proposition 3}\label{app:Proof of proposition 3}
\begin{figure}[h]
    \centering
    \includegraphics[width=1\linewidth]{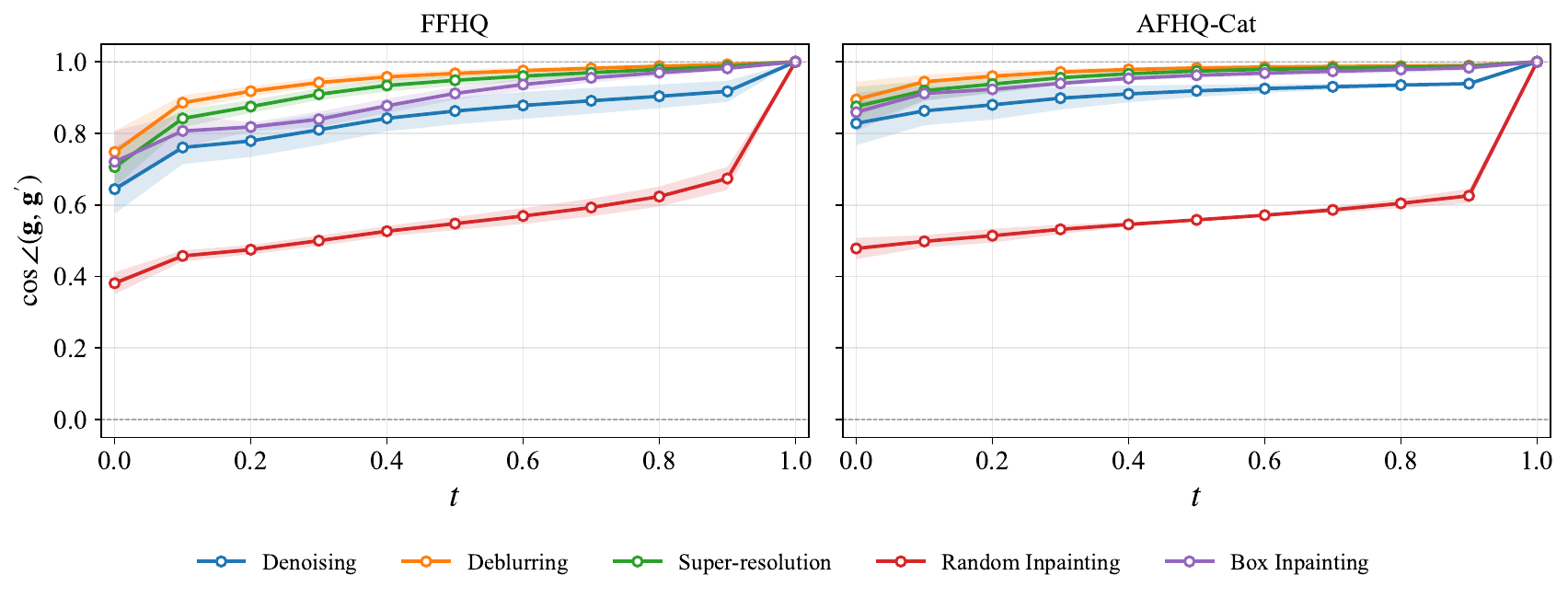}
    \caption{Empirical validation of gradient alignment. Cosine similarity between the true gradient $\mathbf{g}$ and the proposed proxy $\mathbf{g}'$ over the time interval $t \in [0, 1]$. The consistent positive alignment across diverse inverse problems on FFHQ and AFHQ-Cat supports the directional integrity of our proxy-gradient strategy.}
    \label{fig:grad_cosine}
\end{figure}

We consider the composite objective $J(\mathbf{x}_0) = \mathcal{L}(\psi(\mathbf{x}_0), \mathbf{y})$. By the chain rule, the true gradient with respect to the source point $\mathbf{x}_0$ is given by $\mathbf{g} = \nabla_{\mathbf{x}_0} J = \mathbf{M}_N^\top \mathbf{u}$, where $\mathbf{M}_N = \nabla_{\mathbf{x}_0} \psi$ is the cumulative Jacobian and $\mathbf{u} = \nabla_{\mathbf{x}_1}\mathcal{L}(\mathbf{x}_1, \mathbf{y})$. The proxy-gradient update rule is $\mathbf{x}_0^{k+1} = \mathbf{x}_0^k - \eta \mathbf{g}'$, where the proxy gradient is $\mathbf{g}' = C \mathbf{u}$.
To prove convergence, we analyze the inner product between the true gradient $\mathbf{g}$ and the proxy gradient $\mathbf{g}'$:
\begin{equation}
\langle \mathbf{g}, \mathbf{g}' \rangle = \langle \mathbf{M}_N^\top \mathbf{u}, C\mathbf{u} \rangle = C \mathbf{u}^\top \mathbf{M}_N \mathbf{u}.
\end{equation}
By the directional positive alignment assumption, $\mathbf{u}^\top \mathbf{M}_N \mathbf{u} \ge \alpha \|\mathbf{u}\|^2$ with $\alpha > 0$, we have:
\begin{equation}
\langle \mathbf{g}, \mathbf{g}' \rangle \ge C \alpha \|\mathbf{u}\|^2 > 0.
\end{equation}
Since the inner product is strictly positive, $\mathbf{g}'$ is a valid descent direction. Now, assuming $J(\mathbf{x}_0)$ is $L_J$-smooth, we apply the standard descent lemma:
\begin{equation}
\begin{aligned}
J(\mathbf{x}_0^{k+1}) &\le J(\mathbf{x}_0^k) - \eta \langle \mathbf{g}, \mathbf{g}' \rangle + \frac{L_J}{2} \|\eta \mathbf{g}'\|^2 \\
&\le J(\mathbf{x}_0^k) - \eta C \alpha \|\mathbf{u}\|^2 + \frac{L_J \eta^2 C^2}{2} \|\mathbf{u}\|^2 \\
&= J(\mathbf{x}_0^k) - \eta C \left( \alpha - \frac{L_J \eta C}{2} \right) \|\mathbf{u}\|^2.
\end{aligned}
\end{equation}
For any step size $\eta < \frac{2\alpha}{L_J C}$, the objective $J(\mathbf{x}_0)$ decreases monotonically. Summing this inequality over $k = 0, \dots, K-1$ and assuming $J$ is bounded below by $J^*$, we obtain:
\begin{equation}
\sum_{k=0}^{K-1} \|\mathbf{u}_k\|^2 \le \frac{J(\mathbf{x}_0^0) - J^*}{\eta C \left(\alpha - \frac{L_J \eta C}{2}\right)}.
\end{equation}
As $K \to \infty$, the summability of the squared gradients implies that $\lim_{k\to \infty} \|\nabla_{\mathbf{x}_1}\mathcal{L}\| = 0$, thus confirming convergence to a stationary point.
We provide an empirical validation in Figure~\ref{fig:grad_cosine}, where we visualize the cosine similarity between two gradients.

\section{Algorithm}\label{app:algorithm}

Algorithm 1 summarizes the complete P-Flow framework, while Algorithm 2 details the ODE integration process for the flow-consistent path.
\begin{figure}[h]
    \begin{minipage}{0.48\textwidth}
        \begin{algorithm}[H]
            \caption{P-Flow}
            \label{alg:B_flow}
            \begin{algorithmic}[1]
                \REQUIRE Degraded image $\M y$, function H, rate $\eta$, dim $d$.
                \STATE Initialize $\mathbf{x}_0^0 \sim \mathcal{N}(\mathbf{0}, \mathbf{I})$
                \FOR{$k = 0 \rightarrow K-1$}
                    \STATE $\M x^k_1 = \psi(\M x_0^k,N)$
                    \STATE $\mathbf{\tilde{x}}_0^{k+1} = \mathbf{x}_0^k - \eta C\nabla_{\mathbf{x}_1} \mathcal{L}(\mathbf{x}_1, \M y)$
                    \STATE $\mathbf{x}_0^{k+1} \leftarrow  \frac{\sqrt{d}}{\|\mathbf{\tilde{x}}_0^{k+1}\|}\mathbf{\tilde{x}}_0^{k+1}$
                \ENDFOR
                \RETURN $\M x_1 = \psi(\M x_0^K,N)$
            \end{algorithmic}
        \end{algorithm}
    \end{minipage}
    \hfill 
    \begin{minipage}{0.48\textwidth}
        \begin{algorithm}[H]
            \caption{Flow $\psi(\M x_0, N)$}
            \label{alg:FlowForward}
            \begin{algorithmic}[1]
                \REQUIRE Gaussian $\M x_0 \sim \mathcal{N}(\mathbf{0}, \mathbf{I})$, step $N$, field $\M v^\theta$.
                \STATE $t=0$
                \STATE $\delta t = 1/N$
                \FOR{$i=0 \rightarrow N-1$}
                    \STATE $\M x_{t+ \delta t} = \M x_t + \M v_t^\theta(\M x_t)\delta t$
                    \STATE $t = t + \delta t$ \vphantom{$\frac{\sqrt{d}}{\|\mathbf{x}_0^{k+1}\|}$}
                \ENDFOR
                \RETURN $\M x_t$
            \end{algorithmic}
        \end{algorithm}
    \end{minipage}
\end{figure}

\section{Ablation studies and additional results}\label{app:Ablation studies and additional results}
\subsection{Gaussian spherical projection}\label{app:Gaussian spherical projection}
In Table~\ref{tab:ablation_projection}, we evaluate the impact of the Gaussian spherical projection (Step 3) with configurations in Appendix~\ref{app:Implemention details}. The results show that omitting this constraint leads to a consistent degradation in performance across all restoration tasks except deblurring. This empirically confirms that Gaussian spherical projection is indispensable for stabilizing the optimization process and maintaining latent-space consistency within the generative prior. Conversely, deblurring performance shows no decline and even improves slightly when the projection step is removed. This is due to the low-pass properties of the blur operator, which produce smooth gradients that facilitate stable updates of $\mathbf{x}_0$. Consequently, the source point naturally remains within the high-probability manifold of the prior, rendering the projection constraint redundant in this specific regime.

\begin{table}[h]
    \centering
    \caption{Ablation study on the effect of Gaussian spherical projection. We compare the performance of P-Flow with and without the projection step across all tasks and datasets.}
    \label{tab:ablation_projection}

    \begin{subtable}{1\textwidth}
        \centering
        \caption{FFHQ dataset}
        \resizebox{\textwidth}{!}{
            \begin{tabular}{lccccccccccccccc}
                \toprule
                \multirow{2.5}{*}{P-Flow}
                  & \multicolumn{3}{c}{Denoising}
                  & \multicolumn{3}{c}{Deblurring}
                  & \multicolumn{3}{c}{Super-resolution}
                  & \multicolumn{3}{c}{Random inpainting}
                  & \multicolumn{3}{c}{Box inpainting} \\
                \cmidrule(lr){2-4} \cmidrule(lr){5-7} \cmidrule(lr){8-10} \cmidrule(lr){11-13} \cmidrule(lr){14-16}
                  & PSNR  & SSIM  & LPIPS  & PSNR  & SSIM  & LPIPS  & PSNR  & SSIM  & LPIPS  & PSNR  & SSIM  & LPIPS  & PSNR  & SSIM  & LPIPS  \\
                \midrule
                No projection & 20.38 & 0.300 & 0.484 & 27.80 & 0.786 & \textbf{0.140} & 22.80 & 0.542 & 0.265 & 29.66 & 0.847 & 0.069 & 26.80 & 0.752 & 0.102 \\
                With projection  & \textbf{32.37} & \textbf{0.893} & \textbf{0.079} & \textbf{27.87} & \textbf{0.787} & 0.148 & \textbf{27.38} & \textbf{0.816} & \textbf{0.132} & \textbf{32.95} & \textbf{0.922} & \textbf{0.049} & \textbf{30.30} & \textbf{0.930} & \textbf{0.055} \\
                \bottomrule
            \end{tabular}
        }
    \end{subtable}

    \vspace{1.5em}

    \begin{subtable}{1\textwidth}
        \centering
        \caption{AFHQ-Cat dataset}
        \resizebox{\textwidth}{!}{
            \begin{tabular}{lccccccccccccccc}
                \toprule
                \multirow{2.5}{*}{P-Flow}
                  & \multicolumn{3}{c}{Denoising}
                  & \multicolumn{3}{c}{Deblurring}
                  & \multicolumn{3}{c}{Super-resolution}
                  & \multicolumn{3}{c}{Random inpainting}
                  & \multicolumn{3}{c}{Box inpainting} \\
                \cmidrule(lr){2-4} \cmidrule(lr){5-7} \cmidrule(lr){8-10} \cmidrule(lr){11-13} \cmidrule(lr){14-16}
                  & PSNR  & SSIM  & LPIPS  & PSNR  & SSIM  & LPIPS  & PSNR  & SSIM  & LPIPS  & PSNR  & SSIM  & LPIPS  & PSNR  & SSIM  & LPIPS  \\
                \midrule
                No projection & 20.09 & 0.307 & 0.510 & \textbf{28.03} & \textbf{0.750} & \textbf{0.159} & 24.08 & 0.550 & 0.249 & 29.76 & 0.828 & 0.076 & 24.32 & 0.746 & 0.120 \\
                With projection  & \textbf{32.09} & \textbf{0.880} & \textbf{0.106} & 28.01 & 0.741 & 0.165 & \textbf{28.16} & \textbf{0.777} & \textbf{0.157} & \textbf{32.73} & \textbf{0.900} & \textbf{0.060} & \textbf{27.56} & \textbf{0.908} & \textbf{0.081} \\
                \bottomrule
            \end{tabular}
        }
    \end{subtable}
\end{table}

\subsection{Impact of ODE steps}\label{app:Impact of ODE steps}
To investigate the sensitivity of P-Flow to the discretization of the ODE trajectory, we evaluate its performance by varying the number of integration steps $N \in \{3, 5, 8, 10, 15\}$. Other configurations are consistent within Appendix~\ref{app:Implemention details}. Table~\ref{tab:ablation_ode_steps} shows the results for both FFHQ and AFHQ-Cat datasets across five restoration tasks. 
\par
We observe that reconstruction quality do not monotonically improve with $N$ and often saturate or slightly decline at higher values. In contrast, perceptual quality generally benefits from more integration steps, as they facilitate richer detail recovery. Balancing restoration performance and computational overhead, $N \in [5, 10]$ provides a favorable trade-off for practical applications.

\begin{table}[h]
    \centering
    \caption{Ablation study of the number of ODE steps ($N$) on P-Flow performance across various restoration tasks. The optimization iterations are fixed at $K=100$. (a) Results on the FFHQ dataset. (b) Results on the AFHQ-Cat dataset.}
    \label{tab:ablation_ode_steps}

    \begin{subtable}{1\textwidth}
        \centering
        \caption{FFHQ dataset}
        \resizebox{\textwidth}{!}{
            \begin{tabular}{lccccccccccccccc}
                \toprule
                \multirow{2.5}{*}{$N$}
                  & \multicolumn{3}{c}{Denoising}
                  & \multicolumn{3}{c}{Deblurring}
                  & \multicolumn{3}{c}{Super-resolution}
                  & \multicolumn{3}{c}{Random inpainting}
                  & \multicolumn{3}{c}{Box inpainting} \\
                \cmidrule(lr){2-4}\cmidrule(lr){5-7}\cmidrule(lr){8-10}\cmidrule(lr){11-13}\cmidrule(lr){14-16}
                & PSNR & SSIM & LPIPS & PSNR & SSIM & LPIPS & PSNR & SSIM & LPIPS & PSNR & SSIM & LPIPS & PSNR & SSIM & LPIPS \\
                \midrule
                3  & 31.64 & 0.882 & 0.092 & \underline{27.80} & 0.781 & 0.165 & 26.06 & 0.786 & 0.157 & 32.27 & 0.905 & 0.057 & 29.13 & 0.912 & 0.073 \\
                5  & 32.10 & \underline{0.890} & 0.086 & \textbf{27.84} & \textbf{0.787} & 0.144 & 26.96 & 0.807 & 0.144 & 31.30 & 0.884 & 0.059 & 29.77 & \underline{0.925} & 0.064 \\
                8  & \underline{32.33} & \textbf{0.893} & 0.081 & 27.69 & \underline{0.784} & 0.136 & 27.32 & \underline{0.815} & 0.135 & \textbf{32.87} & \textbf{0.926} & 0.043 & \underline{30.19} & \textbf{0.930} & 0.057  \\
                10 & \textbf{32.37} & \textbf{0.893} & \underline{0.079} & 27.56 & 0.780 & \underline{0.131} & \textbf{27.38} & \textbf{0.816} & \underline{0.132} & \underline{32.68} & \underline{0.925} & \underline{0.041} & \textbf{30.30} & \textbf{0.930} & \underline{0.055} \\
                15 & 31.66 & 0.888 & \textbf{0.080} & 27.36 & 0.775 & \textbf{0.123} & \underline{27.37} & \textbf{0.816} & \textbf{0.128} & 32.20 & 0.922 & \textbf{0.039} & 30.11 & 0.924 & \textbf{0.053} \\
                \bottomrule
            \end{tabular}
        }
    \end{subtable}

    \vspace{1.2em}

    \begin{subtable}{1\textwidth}
        \centering
        \caption{AFHQ-Cat dataset}
        \resizebox{\textwidth}{!}{
            \begin{tabular}{lccccccccccccccc}
                \toprule
                \multirow{2.5}{*}{$N$}
                  & \multicolumn{3}{c}{Denoising}
                  & \multicolumn{3}{c}{Deblurring}
                  & \multicolumn{3}{c}{Super-resolution}
                  & \multicolumn{3}{c}{Random inpainting}
                  & \multicolumn{3}{c}{Box inpainting} \\
                \cmidrule(lr){2-4}\cmidrule(lr){5-7}\cmidrule(lr){8-10}\cmidrule(lr){11-13}\cmidrule(lr){14-16}
                & PSNR & SSIM & LPIPS & PSNR & SSIM & LPIPS & PSNR & SSIM & LPIPS & PSNR & SSIM & LPIPS & PSNR & SSIM & LPIPS \\
                \midrule
                3  & 31.21 & 0.851 & 0.127 & \textbf{28.06} & 0.728 & 0.185 & 27.09 & 0.732 & 0.192 & 31.56 & 0.836 & 0.091 & 26.61 & 0.847 & 0.117 \\
                5  & 31.76 & 0.869 & 0.116 & \underline{28.02} & \textbf{0.745} & 0.162 & 27.78 & 0.760 & 0.174 & 32.40 & 0.874 & 0.075 & 27.04 & 0.885 & 0.092 \\
                8  & 32.01 & 0.878 & 0.109 & 27.76 & \underline{0.740} & 0.150 & \underline{28.08} & \underline{0.773} & 0.162 & \underline{32.72} & 0.894 & 0.063 & \underline{27.28} & \underline{0.902} & \underline{0.081} \\
                10 & \underline{32.09} & \underline{0.880} & \underline{0.106} & 27.58 & 0.737 & \underline{0.144} & 28.16 & 0.777 & \underline{0.157} & \textbf{32.73} & \underline{0.900} & \underline{0.060} & \textbf{27.56} & 0.908 & \underline{0.081} \\
                15 & \textbf{32.13} & \textbf{0.883} & \textbf{0.101} & 27.06 & 0.727 & \textbf{0.139} & \textbf{28.22} & \textbf{0.782} & \textbf{0.151} & 32.31 & \textbf{0.903} & \textbf{0.055} & 27.35 & \textbf{0.909} & \textbf{0.080} \\
                \bottomrule
            \end{tabular}
        }
    \end{subtable}
\end{table}

\subsection{Impact of iteration numbers}\label{Impact of interation numbers}
\begin{figure}
    \centering
    \includegraphics[width=1.0\linewidth]{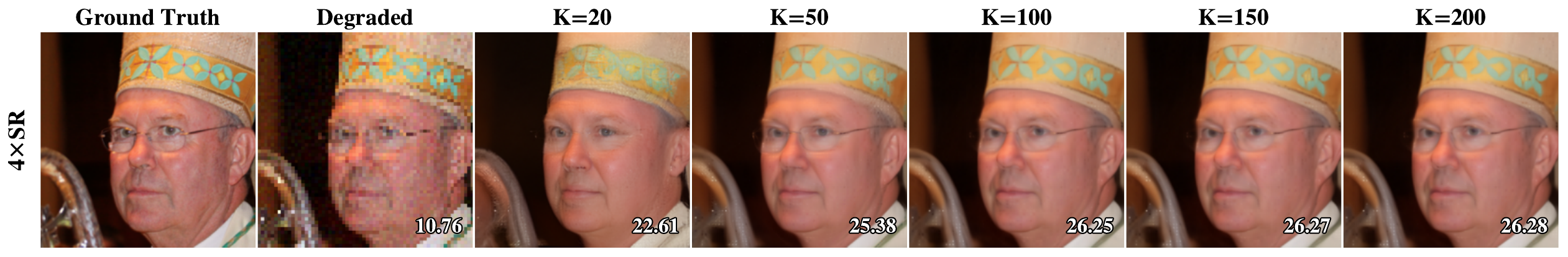}
    \caption{Visualization of $4\times \text{SR}$ results on the FFHQ dataset with parameter $K$.}
    \label{fig:Visualization of K}
\end{figure}

To investigate the sensitivity of P-Flow to the number of optimization iterations $K$, we evaluate its performance by varying $K \in \{20, 50, 100, 150, 200\}$. Other configurations are consistent within Appendix~\ref{app:Implemention details}.  Table~\ref{tab:ablation_optimization_steps} shows the results for both FFHQ and AFHQ-Cat datasets across various restoration tasks, Figure~\ref{fig:Visualization of K} provides visualization of $4\times \text{SR}$ results on the FFHQ dataset.
\par
We observe that the performance does not monotonically improve with the increase of optimization iterations. Notably, while an insufficient number of iterations leads to a significant decline in restoration quality, excessively increasing $K$ does not consistently yield further gains and may even lead to slight fluctuations. This reflects the classic `over-optimization' phenomenon commonly observed in zero-shot generative solvers for inverse problems. In most cases, the optimal range is $K \in [100, 150]$, which provides a favorable balance between restoration accuracy and computational efficiency.

\begin{table}[h]
    \centering
    \caption{Ablation study of the number of optimization steps ($K$) on P-Flow performance across various restoration tasks. (a) Results on the FFHQ dataset. (b) Results on the AFHQ-Cat dataset.}
    \label{tab:ablation_optimization_steps}

    \begin{subtable}{1\textwidth}
        \centering
        \caption{FFHQ dataset}
        \resizebox{\textwidth}{!}{
            \begin{tabular}{lccccccccccccccc}
                \toprule
                \multirow{2.5}{*}{$K$}
                  & \multicolumn{3}{c}{Denoising}
                  & \multicolumn{3}{c}{Deblurring}
                  & \multicolumn{3}{c}{Super-resolution}
                  & \multicolumn{3}{c}{Random inpainting}
                  & \multicolumn{3}{c}{Box inpainting} \\
                \cmidrule(lr){2-4}\cmidrule(lr){5-7}\cmidrule(lr){8-10}\cmidrule(lr){11-13}\cmidrule(lr){14-16}
                & PSNR & SSIM & LPIPS & PSNR & SSIM & LPIPS & PSNR & SSIM & LPIPS & PSNR & SSIM & LPIPS & PSNR & SSIM & LPIPS \\
                \midrule
                20  & 15.88 & 0.621 & 0.305 & 15.55 & 0.559 & 0.314 & 22.50 & 0.695 & 0.190 & 21.79 & 0.776 & 0.135 & 13.35 & 0.516 & 0.380 \\
                50  & \underline{30.93} & \underline{0.883} & \underline{0.087} & 24.60 & 0.751 & 0.172 & 26.28 & 0.792 & 0.136 & 29.72 & \underline{0.897} & \underline{0.053} & 25.21 & 0.868 & 0.112 \\
                100 & \textbf{32.37} & \textbf{0.893} & \textbf{0.079} & \underline{27.87} & \textbf{0.787} & \textbf{0.148} & 27.38 & \underline{0.816} & \textbf{0.132} & \underline{32.95} & \textbf{0.921} & \textbf{0.049} & \textbf{30.29} & \textbf{0.930} & \textbf{0.055}  \\
                150 & 29.24 & 0.872 & 0.091 & \textbf{28.09} & \underline{0.783} & \underline{0.150} & \underline{27.46} & \textbf{0.818} & \underline{0.133} & 33.18 & \textbf{0.921} & \underline{0.053} & \underline{28.52} & \underline{0.911} & \underline{0.067} \\
                200 & 24.28 & 0.839 & 0.107 & 27.97 & 0.769 & 0.157 & \textbf{27.47} & \textbf{0.818} & \underline{0.133} & \textbf{33.22} & \textbf{0.921} & \underline{0.053} & 22.86 & 0.844 & 0.110\\
                \bottomrule
            \end{tabular}
        }
    \end{subtable}

    \vspace{1.2em}

    \begin{subtable}{1\textwidth}
        \centering
        \caption{AFHQ-Cat dataset}
        \resizebox{\textwidth}{!}{
            \begin{tabular}{lccccccccccccccc}
                \toprule
                \multirow{2.5}{*}{$K$}
                  & \multicolumn{3}{c}{Denoising}
                  & \multicolumn{3}{c}{Deblurring}
                  & \multicolumn{3}{c}{Super-resolution}
                  & \multicolumn{3}{c}{Random inpainting}
                  & \multicolumn{3}{c}{Box inpainting} \\
                \cmidrule(lr){2-4}\cmidrule(lr){5-7}\cmidrule(lr){8-10}\cmidrule(lr){11-13}\cmidrule(lr){14-16}
                & PSNR & SSIM & LPIPS & PSNR & SSIM & LPIPS & PSNR & SSIM & LPIPS & PSNR & SSIM & LPIPS & PSNR & SSIM & LPIPS \\
                \midrule
                20  & 16.28 & 0.662 & 0.303 & 14.55 & 0.518 & 0.364 & 23.09 & 0.661 & 0.195 & 19.26 & 0.720 & 0.204 & 14.36 & 0.594 & 0.346 \\
                50  & 29.90 & 0.869 & 0.118 & 23.80 & 0.700 & 0.197 & 27.29 & \underline{0.765} & \textbf{0.150} & 27.97 & 0.863 & 0.081 & 23.82 & 0.858 & 0.129 \\
                100 & \textbf{32.09} & \textbf{0.880} & \textbf{0.106} & \underline{28.01} & \textbf{0.741} & \textbf{0.165} & \underline{28.16} & \textbf{0.777} & \underline{0.157} & 32.63 & \underline{0.897} & \textbf{0.059} & \underline{27.40} & \underline{0.908} & \textbf{0.079} \\
                150 & \underline{32.01} & \textbf{0.880} & \textbf{0.106} & \textbf{28.24} & \underline{0.735} & \underline{0.166} & \textbf{28.17} & \textbf{0.777} & 0.161 & \underline{33.03} & \textbf{0.899} & \underline{0.065} & \textbf{27.46} & \textbf{0.909} & \underline{0.082} \\
                200 & 31.24 & \underline{0.878} & \underline{0.109} & 28.04 & 0.720 & 0.173 & \textbf{28.17} & \textbf{0.777} & 0.162 & \textbf{33.07} & 0.896 & 0.067 & 26.79 & 0.904 & 0.086 \\
                \bottomrule
            \end{tabular}
        }
    \end{subtable}
\end{table}

\subsection{Impact of learning rate}\label{app:learning rate}
To investigate the sensitivity of P-Flow to the learning rate, we evaluate its performance by varying the learning rate $\eta \in \{0.01,0.02,0.05,0.1,0.2,0.5\}$. Other configurations are consistent within Appendix~\ref{app:Implemention details}. Table~\ref{tab:ablation_learning_rate} shows the results for both FFHQ and AFHQ-Cat datasets across five restoration tasks. 
\par
Notably, while an small learning rate leads to insufficient convergence and a decline in restoration quality, an overly large learning rate may cause convergence collapse or significant numerical instability. In most cases, the optimal range for the learning rate is $[0.1, 0.2]$, which provides a favorable balance between restoration accuracy and optimization stability.

\begin{table}[t]
    \centering
    \caption{Ablation study of the learning rate ($\eta$) on P-Flow performance across various restoration tasks. (a) Results on the FFHQ dataset. (b) Results on the AFHQ-Cat dataset.}
    \label{tab:ablation_learning_rate}

    \begin{subtable}{1\textwidth}
        \centering
        \caption{FFHQ dataset}
        \resizebox{\textwidth}{!}{
            \begin{tabular}{lccccccccccccccc}
                \toprule
                \multirow{2.5}{*}{$\eta$}
                  & \multicolumn{3}{c}{Denoising}
                  & \multicolumn{3}{c}{Deblurring}
                  & \multicolumn{3}{c}{Super-resolution}
                  & \multicolumn{3}{c}{Random inpainting}
                  & \multicolumn{3}{c}{Box inpainting} \\
                \cmidrule(lr){2-4}\cmidrule(lr){5-7}\cmidrule(lr){8-10}\cmidrule(lr){11-13}\cmidrule(lr){14-16}
                & PSNR & SSIM & LPIPS & PSNR & SSIM & LPIPS & PSNR & SSIM & LPIPS & PSNR & SSIM & LPIPS & PSNR & SSIM & LPIPS \\
                \midrule
                0.01 & 30.61 & 0.866 & 0.098 & 26.65 & 0.753 & 0.173 & 19.51 & 0.586 & 0.292 & 26.94 & 0.798 & 0.132 & 27.99 & 0.887 & 0.080   \\
                0.02 & 31.54 & 0.884 & 0.087 & 27.28 & 0.772 & 0.160 & 22.38 & 0.663 & 0.218 & 29.48 & 0.858 & 0.074 & 29.08 & 0.917 & 0.056   \\
                0.05 & 31.94 & 0.890 & 0.087 & 27.74 & 0.784 & 0.152 & 24.63 & 0.738 & 0.162 & 31.35 & 0.901 & \underline{0.046} & 29.99 & \underline{0.929} & \textbf{0.050}    \\
                0.1  & \underline{32.17} & \underline{0.893} & \underline{0.083} & \underline{27.85} & \underline{0.786} & \underline{0.150} & 26.08 & 0.784 & \underline{0.139} & 32.34 & 0.916 & \textbf{0.043} & \underline{30.23} & \textbf{0.930} & \underline{0.052}   \\
                0.2  & \textbf{32.34} & \textbf{0.894} & \textbf{0.080} & \textbf{28.07} & \textbf{0.788} & \textbf{0.143} & \underline{27.12} & \underline{0.810} & \textbf{0.132} & \underline{32.88} & \textbf{0.921} & 0.047 & \textbf{30.27} & \textbf{0.930} & 0.053   \\
                0.5  & 21.67 & 0.768 & 0.149 & 26.49 & 0.765 & 0.152 & \textbf{27.64} & \textbf{0.819} & \textbf{0.132} & \textbf{32.99} & \underline{0.920} & 0.051 & 24.06 & 0.854 & 0.105  \\
                \bottomrule
            \end{tabular}
        }
    \end{subtable}

    \vspace{1.2em}

    \begin{subtable}{1\textwidth}
        \centering
        \caption{AFHQ-Cat dataset}
        \resizebox{\textwidth}{!}{
            \begin{tabular}{lccccccccccccccc}
                \toprule
                \multirow{2.5}{*}{$\eta$}
                  & \multicolumn{3}{c}{Denoising}
                  & \multicolumn{3}{c}{Deblurring}
                  & \multicolumn{3}{c}{Super-resolution}
                  & \multicolumn{3}{c}{Random inpainting}
                  & \multicolumn{3}{c}{Box inpainting} \\
                \cmidrule(lr){2-4}\cmidrule(lr){5-7}\cmidrule(lr){8-10}\cmidrule(lr){11-13}\cmidrule(lr){14-16}
                & PSNR & SSIM & LPIPS & PSNR & SSIM & LPIPS & PSNR & SSIM & LPIPS & PSNR & SSIM & LPIPS & PSNR & SSIM & LPIPS \\
                \midrule
                0.01 & 30.53 & 0.845 & 0.125 & 27.10 & 0.710 & 0.184 & 20.59 & 0.566 & 0.297 & 27.69 & 0.773 & 0.137 & 25.49 & 0.860 & 0.106   \\
                0.02 & 31.46 & 0.868 & 0.111 & 27.64 & 0.728 & 0.173 & 23.58 & 0.642 & 0.213 & 29.89 & 0.836 & 0.079 & 26.41 & 0.894 & \underline{0.081}   \\
                0.05 & 31.83 & 0.875 & 0.112 & \underline{27.98} & 0.737 & 0.167 & 25.84 & 0.712 & 0.159 & 31.57 & 0.880 & \underline{0.054} & 27.38 & 0.905 & \textbf{0.075}    \\
                0.1  & \underline{31.97} & \underline{0.878} & \underline{0.109} & \underline{27.98} & \underline{0.738} & \underline{0.166} & 27.26 & 0.760 & \textbf{0.150} & 32.38 & 0.896 & \textbf{0.052} & \underline{27.54} & \underline{0.907} & \textbf{0.075}   \\
                0.2  & \textbf{32.09} & \textbf{0.880} & \textbf{0.106} & \textbf{28.01} & \textbf{0.741} & \textbf{0.165} & \underline{28.07} & \underline{0.776} & \underline{0.154} & \textbf{32.61} & \textbf{0.899} & 0.056 & \textbf{27.56} & \textbf{0.908} & \underline{0.081}   \\
                0.5  & 17.47 & 0.693 & 0.239 & 24.90 & 0.704 & 0.188 & \textbf{28.21} & \textbf{0.777} & 0.159 & \underline{32.60} & \underline{0.898} & 0.064 & 22.00 & 0.832 & 0.153  \\
                \bottomrule
            \end{tabular}
        }
    \end{subtable}
\end{table}

\subsection{Computation time and memory}\label{app:Computation time and memory}
As shown in Table~\ref{tab:benchmark_algorithms_horizontal}, we evaluate the computational time and memory load for several methods on the FFHQ random inpainting task ($256 \times 256$), with configurations in Appendix~\ref{app:Implemention details}. All benchmarks are conducted on an NVIDIA H200 SXM 141GB GPU. We report the inference latency and peak GPU memory usage per image, with all results obtained by averaging over 128 test images.
\par
Compared to traditional source-optimization methods like D-Flow, P-Flow achieves a \textbf{65.9$\times$} reduction in peak VRAM (21.09 GB to 0.32 GB) and a \textbf{28.6$\times$} speedup in inference time (149.28s to 5.22s). This drastic reduction empirically validates our complexity analysis in Section~\ref{sec:complexity}, as our proxy-gradient strategy completely eliminates the need to cache long-chain ODE activations. Furthermore, compared to lightweight PnP methods like PnP-Flow, P-Flow maintains an almost identical memory footprint (0.32 GB), while its ODE forwarding nature incurs a practical runtime of under 10 seconds.

\begin{table}[h]
    \centering
    \caption{Time and memory metrics per image on the super-resolution task on FFHQ ($256\times 256$).}
    \label{tab:benchmark_algorithms_horizontal}
    \vspace{1mm} 
    \resizebox{\textwidth}{!}{
    \begin{tabular}{lcccccccc}
        \toprule
        Method &  DiffPIR &  PnP-GS & OT-ODE & Flow-Priors  & PnP-Flow   & Flower    & D-Flow    & P-Flow \\
        \midrule
        Time [s] &  1.55   &   1.27   & 4.60      & 66.58      & 1.32     & 1.16    & 149.28      & 5.22     \\
        Memory [GB] &   0.67  &   8.44   & 2.27     & 15.44     & 0.32     & 0.32     & 21.09      & 0.32 \\
        \bottomrule
    \end{tabular}
    }
\end{table}

\section{Implementation details}\label{app:Implemention details}
The optimal configuration of all method are listed in Table~\ref{tab:params_ffhq} (FFHQ) and Table~\ref{tab:params_afhq} (AFHQ-Cat).
\begin{table}[h]
    \caption{Hyperparameters for all methods on the FFHQ dataset. }
    \vspace{1em}
    \label{tab:params_ffhq}
    \centering
    \resizebox{1\hsize}{!}{
        \begin{tabular}{llccccc}
            \toprule
            \textbf{Method} &  Hyperparameters & {Denoising} & Deblurring & Super-resolution & Random inpainting & Box inpainting \\
            \midrule
            DiffPIR  & $\zeta$ (blending) & 1.0 & 1.0 & 1.0 & 1.0 & N/A \\ 
                     & $\lambda$ (regularization) & 1.0 & 100.0 & 10.0 & 1000.0 & N/A \\ 
            \midrule
            PnP-GS   & $\gamma$ (learning rate) & - & 2.0 & 2.0 & 1.0 & N/A \\ 
                     & $\alpha$ (inertia param.) & 0.3 & 0.3 & 1.0 & 0.3 & N/A \\
                     & $\sigma_f$ (factor for noise input) & 1.5 & 1.8 & 5.0 & 1.0 & N/A \\
                     & $n_\text{iter}$ (number of iter.) & 5 & 60 & 50 & 30 & N/A \\ 
            \midrule
            OT-ODE    & $t_0$ (initial time) & 0.1 & 0.2 & 0.1 & 0.1 & 0.1\\
                     & $\gamma$ & constant& time-dependent & constant & constant & time-dependent\\ 
            \midrule
            Flow-Priors & $\lambda$ (regularization) & 30 & 3,000 & 30,000 & 30,000 & 10,000\\
                       & $\eta$ (learning rate) & 0.01 & 0.01 & 0.1 & 0.01 & 0.01\\ 
            \midrule
            PnP-Flow  & $\alpha$ (learning-rate factor) & 0.8 & 0.01 & 0.01 & 0.01 & 0.3\\
                     & $N$ (Number of time steps) & 100 & 100 & 100 & 100 & 100\\
                     & $N_{\mathrm{Avg}}$ (Number of averagings) & 1 & 1 & 1 & 1 & 1\\
            \midrule
            Flower  & $\gamma$ (refinement uncertainty) & 0 & 0 & 0 & 0 & 0\\
                     & $N$ (Number of time steps) & 100 & 100 & 100 & 100 & 100\\
                     & $N_{\mathrm{Avg}}$ (Number of averagings) & 1 & 1 & 1 & 1 & 1\\
            \midrule
            D-Flow    & $\lambda$ (regularization) & 0.001 & 0.001 & 0.001 & 0.01 & 0.001 \\ 
                     & $\alpha$ (blending) & 0.1 & 0.1 & 0.1 & 0.1 & 0.1 \\
                     & $n_\text{iter}$ (number of iter.) & 3 & 7 & 10 & 20 & 9 \\ 
            \midrule
            P-Flow  & $\eta$ (learning rate) & 0.3 & 0.2 & 0.5 & 0.4 & 0.3\\
         & $K$ (Number of iter.) & 100 & 100 & 100 & 100 & 100\\
         & $N$ (Number of ODE steps) & 10 & 5 & 10 & 5 & 10\\
            \bottomrule
        \end{tabular}    
    }       
\end{table}

\begin{table}[h]
    \caption{
    Hyperparameters for all methods on the AFHQ-Cat dataset. }
    \vspace{1em}
    \label{tab:params_afhq}
    \centering
    \resizebox{1\hsize}{!}{
        \begin{tabular}{llccccc}
            \toprule
            \textbf{Method} &  & {Denoising} & {Deblurring} & {Super-resolution} & {Random inpainting} & {Box inpainting} \\
            \midrule
            DiffPIR & $\zeta$ (blending) & 1.0 & 1.0 & 1.0 & 1.0 & N/A \\ 
                     & $\lambda$ (regularization) & 1.0 & 100.0 & 10.0 & 10.0 & N/A \\ 
            \midrule
            PnP-GS   & $\gamma$ (learning rate) & - & 2.0 & 2.0 & 1.0& N/A \\ 
                     & $\alpha$ (inertia param.) & 1.0 & 0.3 & 1.0 & 0.5 & N/A \\
                     & $\sigma_f$ (factor for noise input) & 1.0 & 1.8 & 5.0 & 1.0 & N/A \\
                     & $n_\text{iter}$ (number of iter.) & 1 & 60 & 50 & 23 & N/A \\ 
            \midrule
            OT-ODE    & $t_0$ (initial time) & 0.1 & 0.3 & 0.1 & 0.1 & 0.1 \\
                     & $\gamma$ &time-dependent &time-dependent  & constant & constant & time-dependent\\ 
            \midrule
            Flow-Priors & $\lambda$ (regularization) & 30 & 1,000 & 30,000 & 30,000 & 10,000 \\
                        & $\eta$ (learning rate) & 0.01 & 0.01 & 0.1 & 0.01 & 0.01 \\ 
            \midrule
            PnP-Flow & $\alpha$ (learning-rate factor) & 0.8 & 0.01 & 0.01 & 0.01 & 0.3 \\
                     & $N$ (Number of time steps) & 100 & 100 & 100 & 100 & 100 \\
                     & $N_{\mathrm{Avg}}$ (Number of averagings) & 1 & 1 & 1 & 1 & 1\\
            \midrule
            Flower  & $\gamma$ (refinement uncertainty) & 0 & 0 & 0 & 0 & 0\\
                     & $N$ (Number of time steps) & 100 & 100 & 100 & 100 & 100\\
                     & $N_{\mathrm{Avg}}$ (Number of averagings) & 1 & 1 & 1 & 1 & 1\\
            \midrule
            D-Flow   & $\lambda$ (regularization) & 0.001 & 0.01 & 0.001 & 0.001 & 0.01 \\ 
            & $\alpha$ (blending) & 0.1 & 0.5 & 0.1 & 0.1 & 0.1 \\
            & $n_\text{iter}$ (number of iter.) & 3 & 20 & 20 & 20 & 9 \\ 
            \midrule
            P-Flow & $\eta$ (learning rate) & 0.2 & 0.2 & 0.3 & 0.3 & 0.2\\
         & $K$ (Number of iter.) & 100 & 100 & 100 & 100 & 100\\
         & $N$ (Number of ODE steps) & 10 & 5 & 10 & 10 & 10\\
            \bottomrule
        \end{tabular}    
    }       
\end{table}

\section{Experiment statistical significance}\label{app:statistical significance}
\begin{figure}
    \centering
    \includegraphics[width=1\linewidth]{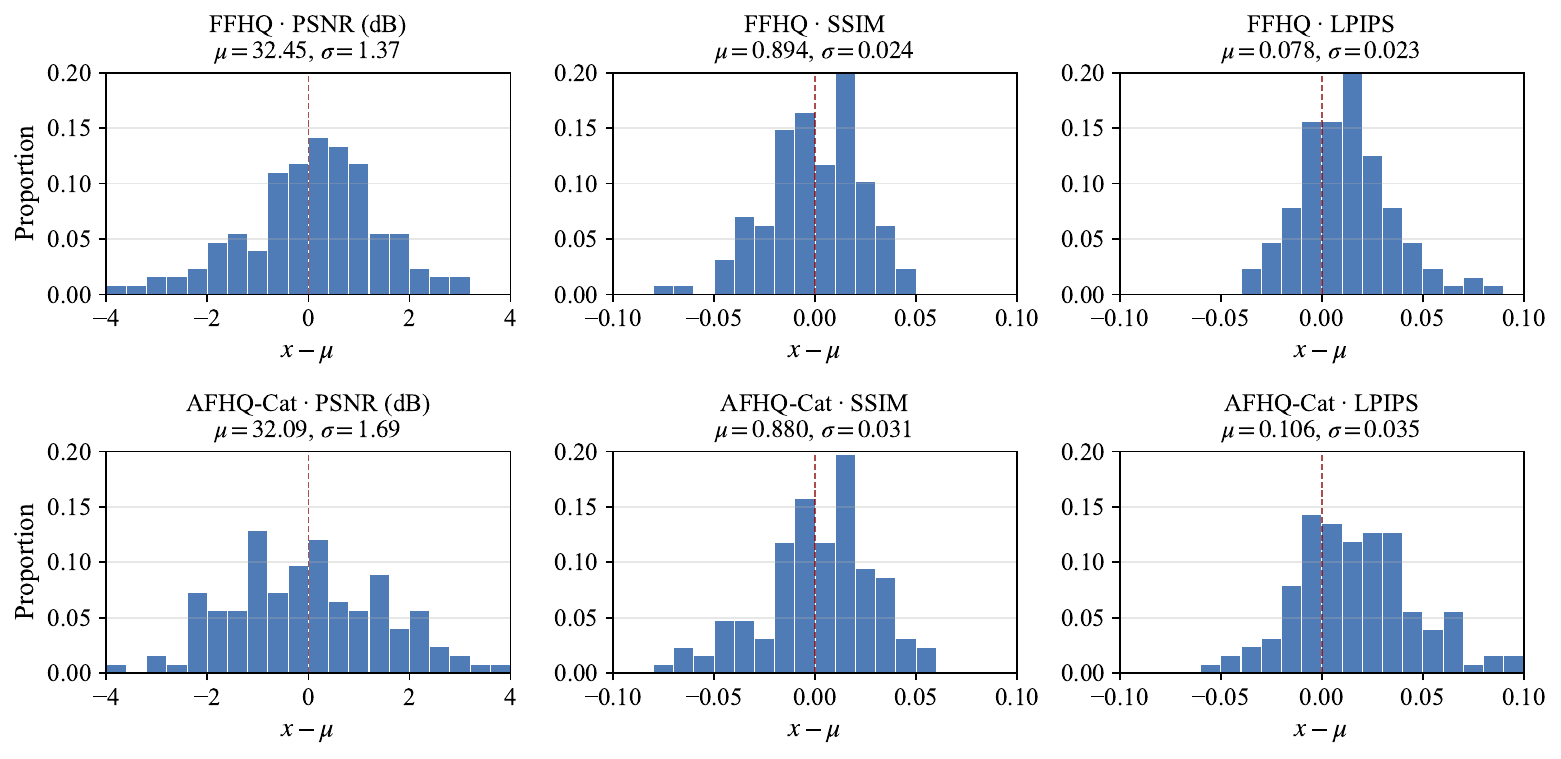}
    \caption{Quantitative error distributions for denoising. The histograms illustrate the variation in PSNR, SSIM, and LPIPS metrics for FFHQ and AFHQ-Cat. Values for mean performance ($\mu$) and standard deviation ($\sigma$) are provided for each metric.}
    \label{fig:sr_distribution}
\end{figure}

Figure~\ref{fig:sr_distribution} illustrates the statistical distribution and standard deviation of the results for the denoising.

\section{Additional results}\label{app:Additional results}
\subsection{Additional cross-task comparisons}\label{app:Additional visual results}
\begin{figure}[h]
    \centering
    \includegraphics[width=0.95\linewidth]{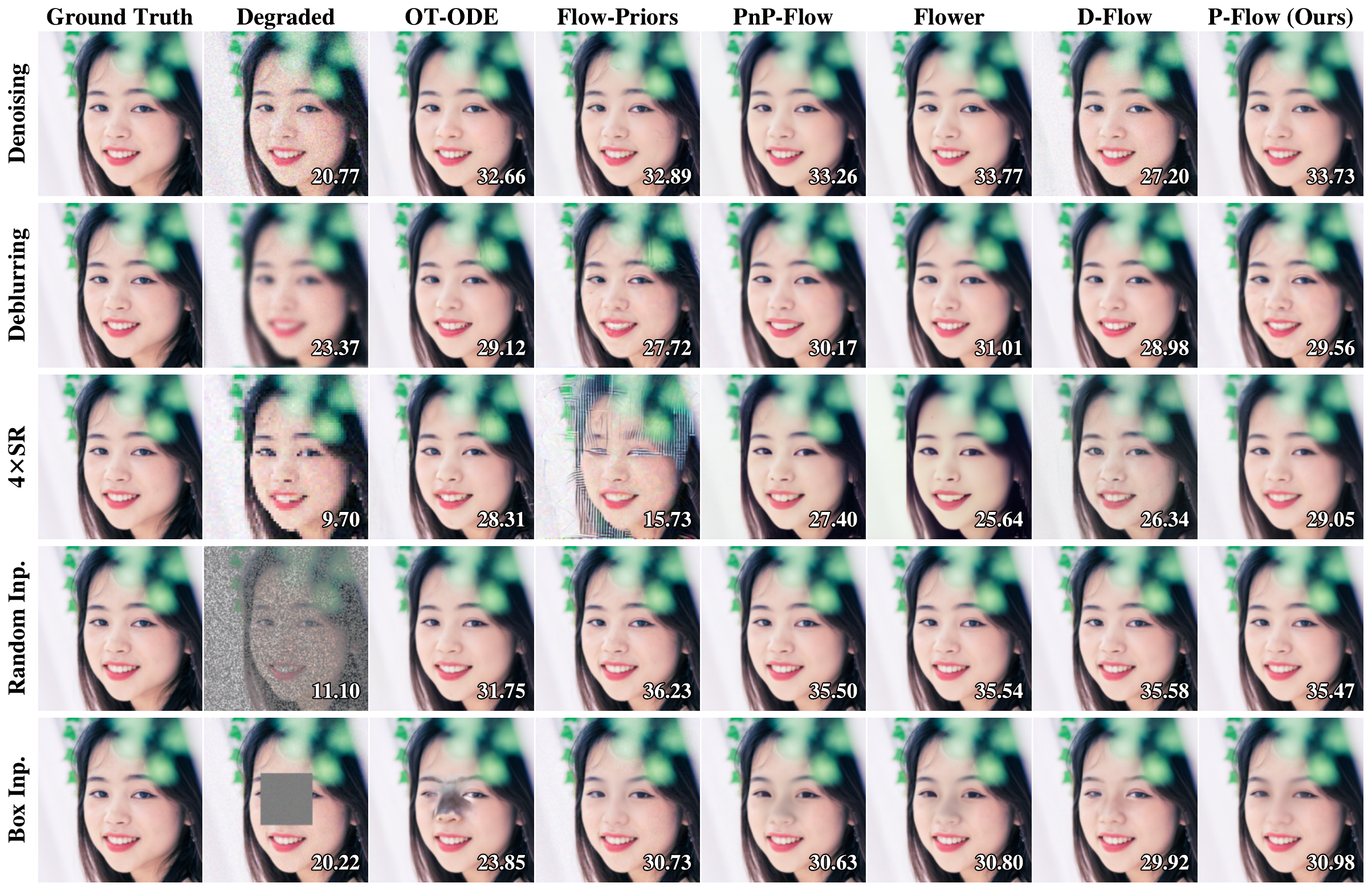}
    \caption{Comparison of image restoration methods on the FFHQ dataset using the PSNR metric.}
    \label{fig:comparison_figure_ffhq_batch26}
\end{figure}

\begin{figure}[h]
    \centering
    \includegraphics[width=0.95\linewidth]{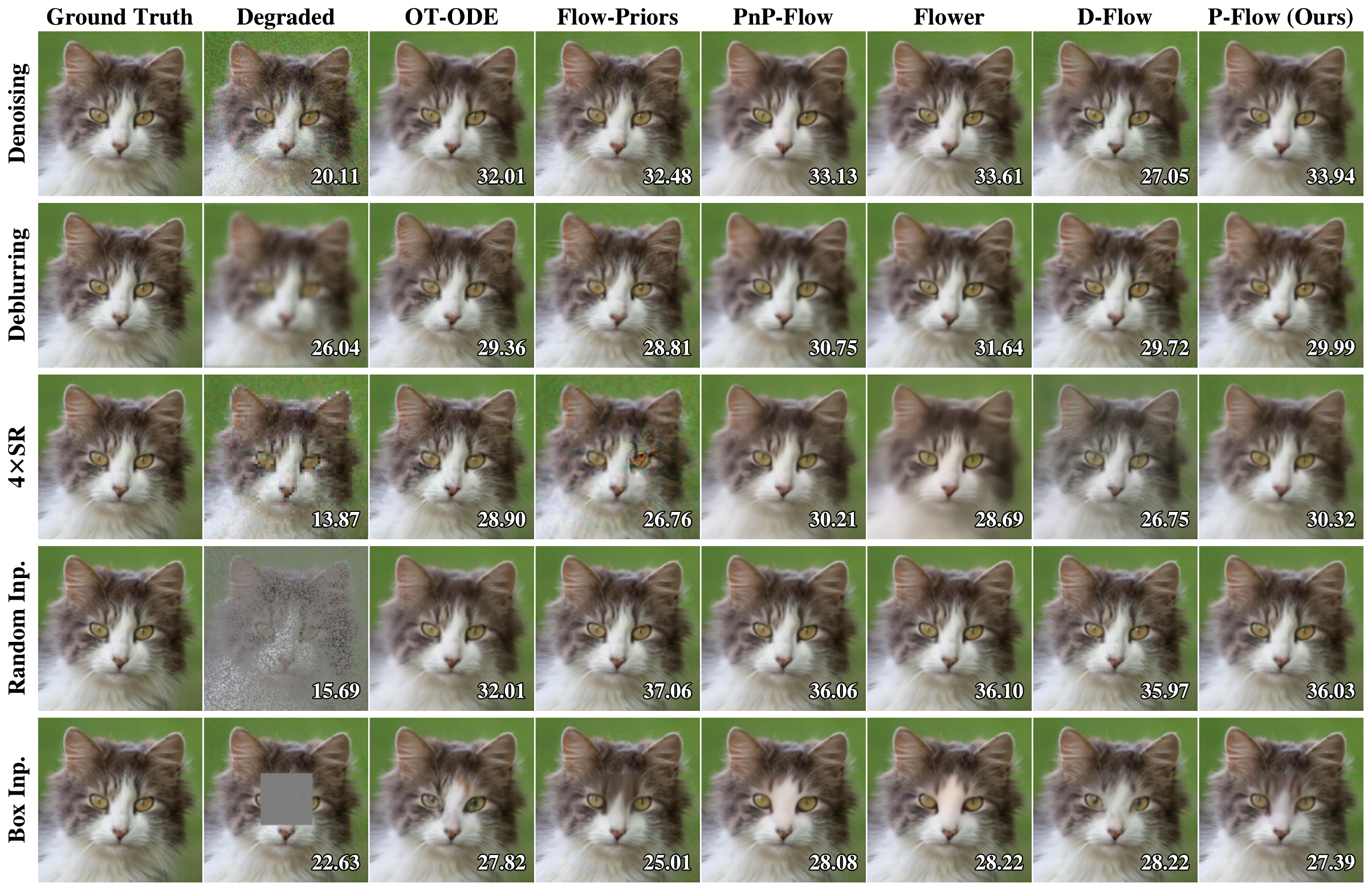}
    \caption{Comparison of image restoration methods on the AFHQ-Cat dataset using the PSNR metric.}
    \label{fig:comparison_figure_afhq_cat_batch9}
\end{figure}

Figure~\ref{fig:comparison_figure_ffhq_batch26} and Figure~\ref{fig:comparison_figure_afhq_cat_batch9} provide additional comparisons on FFHQ and AFHQ-Cat, respectively. Unlike the main-text examples,
the two figures evaluate all restoration tasks on the same underlying image. This
layout highlights cross-task consistency and makes it easier to compare failure
modes across methods, including texture over-smoothing, identity distortion,
color shifts, and structural artifacts. These results complement the quantitative
metrics by showing that P-Flow preserves coherent global structure and realistic
details across different degradation types.

\subsection{Severely ill-posed conditions}\label{app:Severely ill-posed conditions}
\begin{figure}
    \centering
    \includegraphics[width=1.0\linewidth]{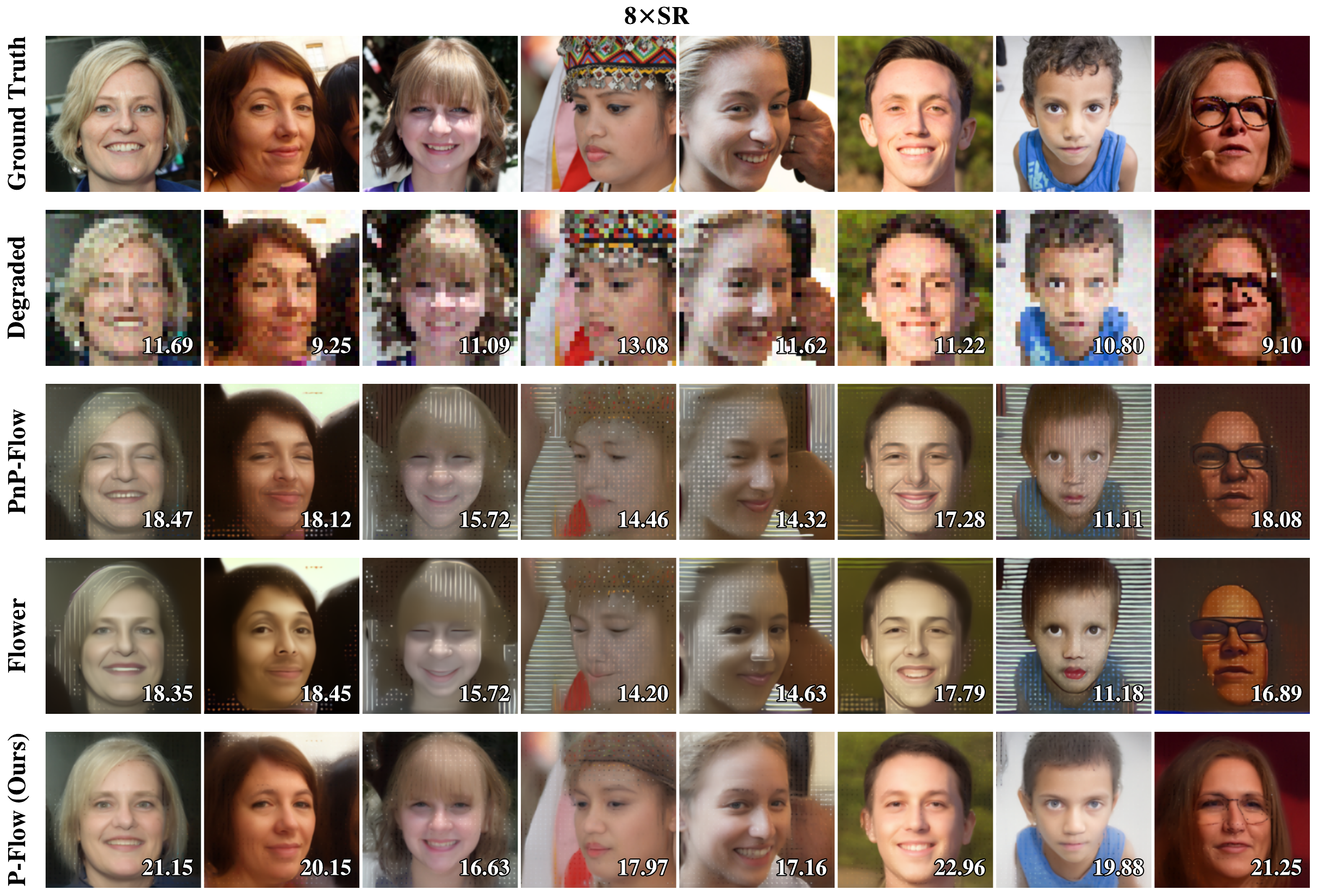}
    \caption{Visual comparison for $8 \times \text{SR}$  on the FFHQ dataset. We compare the proposed P-Flow with PnP-Flow and Flower. The PSNR metric for each restoration is indicated in the bottom-right corner.}
    \label{fig:8xsr}
\end{figure}

\begin{figure}
    \centering
    \includegraphics[width=1.0\linewidth]{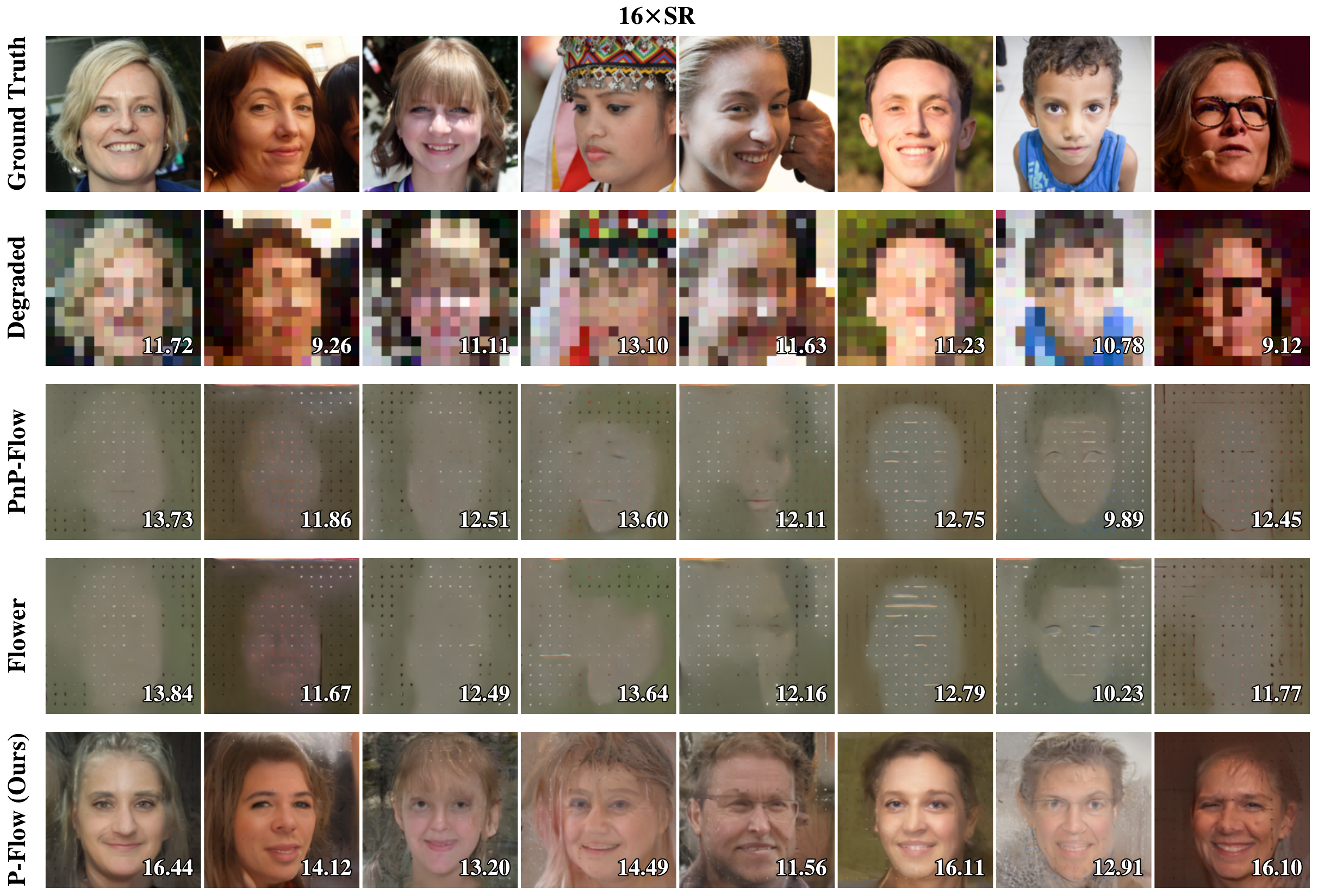}
    \caption{Visual comparison for $16 \times \text{SR}$  on the FFHQ dataset. We compare the proposed P-Flow with PnP-Flow and Flower. The PSNR metric for each restoration is indicated in the bottom-right corner.}
    \label{fig:16xsr}
\end{figure}

\begin{figure}
    \centering
    \includegraphics[width=1.0\linewidth]{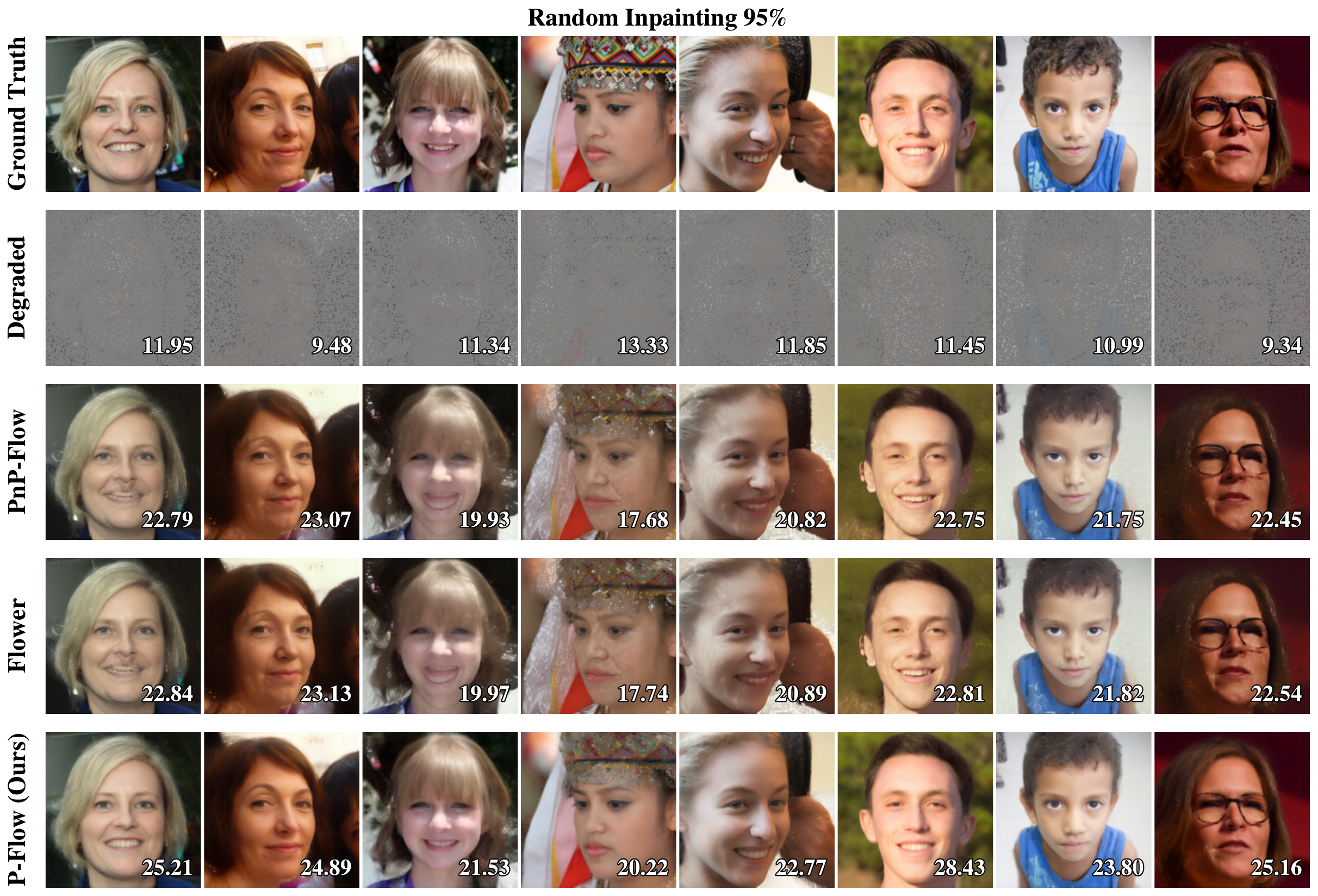}
    \caption{Visual comparison for random inpainting with $95\%$ masked pixels on the FFHQ dataset. We compare the proposed P-Flow with PnP-Flow and Flower. The PSNR metric for each restoration is indicated in the bottom-right corner.}
    \label{fig:inpainting95}
\end{figure}

\begin{figure}
    \centering
    \includegraphics[width=1.0\linewidth]{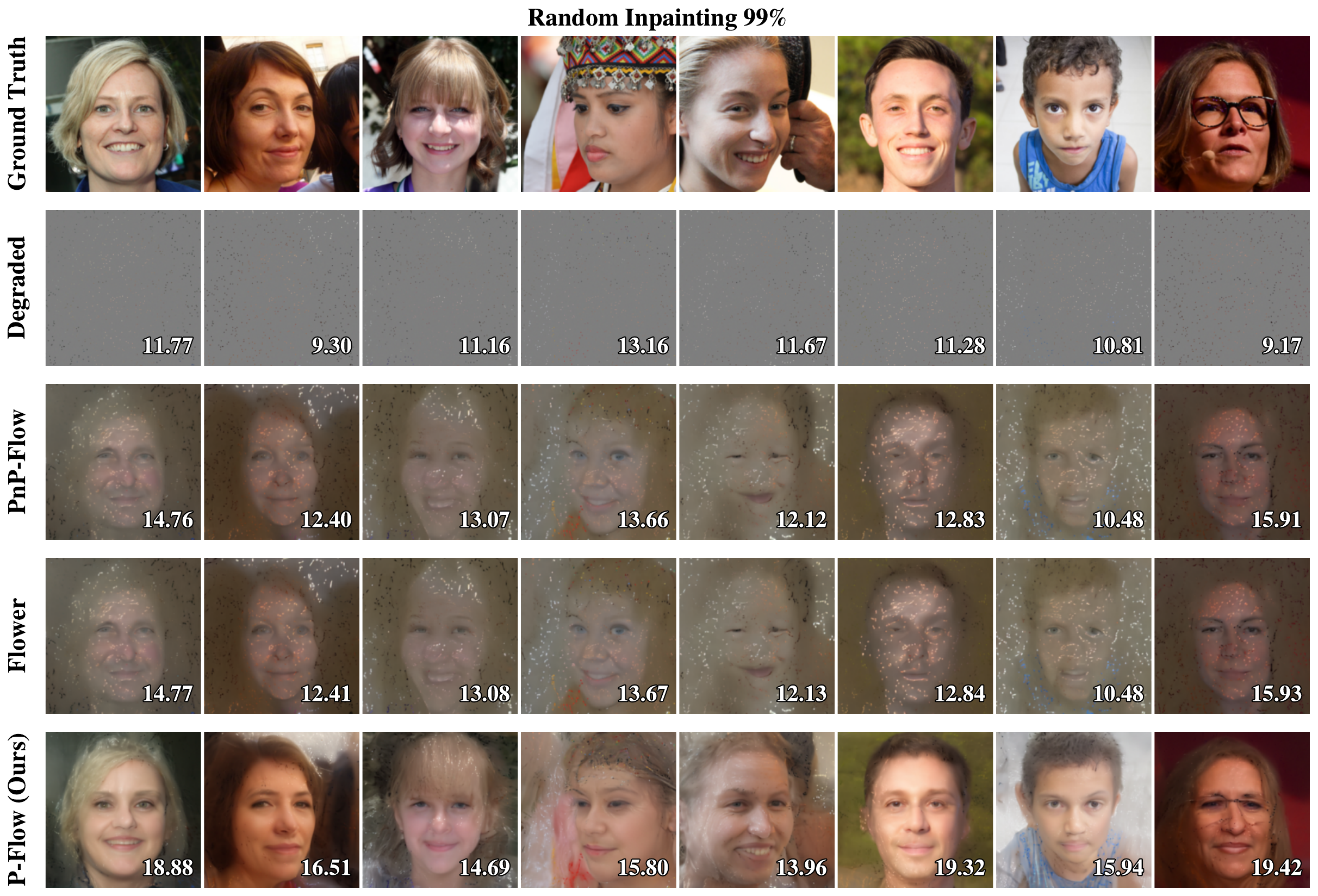}
    \caption{Visual comparison for random inpainting with $99\%$ masked pixels on the FFHQ dataset. We compare the proposed P-Flow with PnP-Flow and Flower. The PSNR metric for each restoration is indicated in the bottom-right corner.}
    \label{fig:inpainting99}
\end{figure}

We further evaluate the stability of P-Flow under extremely ill-posed measurement regimes (Table~\ref{tab:severely ill-posed conditions}), including high-ratio super-resolution ($8\times$ and $16\times$) and extremely sparse random inpainting (95\% and 99\% masking). In the super-resolution tasks (Figure~\ref{fig:8xsr} and Figure~\ref{fig:16xsr}), P-Flow significantly outperforms Flower and PnP-Flow, preserving coherent facial structures where the baselines begin to collapse. In the extreme random inpainting tasks (Figure~\ref{fig:inpainting95} and Figure~\ref{fig:inpainting99}), the baselines suffer from severe identity distortion at 95\% masking and near-total perceptual collapse at 99\% masking. By contrast, P-Flow consistently recovers more coherent global structures and more plausible facial identities.
\par
The underlying reason is that these degradations provide only weak or sparse measurement guidance. In high-ratio super-resolution, most high-frequency information is absent from the observation. In extreme random inpainting, only a tiny fraction of pixels is available to constrain the reconstruction. In both cases, the data-consistency signal is insufficient to determine the missing semantic content, so the reconstruction quality depends heavily on how well the method preserves and exploits the generative prior.
\par
PnP-Flow relies on repeated data-fidelity updates of the intermediate states. Under extremely weak measurements, these updates provide limited information about the missing structures while still perturbing the trajectory. Flower mitigates this issue by refining the predicted destination with a proximal measurement step, but its refinement is still driven by highly incomplete observations. As the measurement becomes too sparse or too low-dimensional, the refined destination can become weakly constrained, leading to overly conservative or structurally unstable reconstructions.
\par
P-Flow avoids repeatedly correcting intermediate states. Instead, it performs optimization in the latent source space and then generates the reconstruction through the original flow trajectory. The Gaussian spherical projection prevents the source variable from drifting away from the typical prior region, while the unmodified ODE trajectory preserves the learned generative dynamics. Therefore, even when the measurement signal is vanishingly weak, P-Flow can rely on the intact generative path to produce semantically coherent and visually plausible structures.

\begin{table}[t]
    \centering
    \caption{Quantitative comparison under severely ill-posed conditions. We evaluate the performance of PnP-Flow, Flower, and P-Flow over an average of 128 test images. (a) Results on the FFHQ dataset. (b) Results on the AFHQ dataset.}
    \label{tab:severely ill-posed conditions}
    
    \begin{subtable}{1\textwidth}
        \centering
        \caption{FFHQ dataset}
        \resizebox{\textwidth}{!}{
            \begin{tabular}{lccccccccccccccc}
                \toprule
                \multirow{2.5}{*}{Method}
                  & \multicolumn{3}{c}{$8\times\text{SR}$}
                  & \multicolumn{3}{c}{$16\times\text{SR}$}
                  & \multicolumn{3}{c}{$95\%$ Random inp.}
                  & \multicolumn{3}{c}{$99\%$ Random inp.} 
                  & \multicolumn{3}{c}{$\sigma=1.0$ Denoising} \\
                \cmidrule(lr){2-4}\cmidrule(lr){5-7}\cmidrule(lr){8-10}\cmidrule(lr){11-13}\cmidrule(lr){14-16}
                & PSNR  & SSIM  & LPIPS  & PSNR  & SSIM  & LPIPS  & PSNR  & SSIM  & LPIPS  & PSNR  & SSIM  & LPIPS  & PSNR  & SSIM  & LPIPS \\
                \midrule 
                Degraded  & 11.35 & 0.323 &  0.974 & 11.36 & 0.359  &  0.896 &  11.59 &  0.255 &  1.108 &  11.50 & 0.307  &  1.032 & 9.16  & 0.038 & 1.281\\
                Flow-Priors & 15.86 &  0.252 &  0.616 & \underline{13.38}  &  0.206 & 0.702  & 19.41  & 0.451  & 0.436    &  \underline{18.49}  & 0.420  & 0.479 &  17.13 & 0.170 & 0.795\\
                PnP-Flow & 15.70  &  0.442 & 0.504 &  13.26 & \underline{0.417}  &  \underline{0.583}  & 22.21  & 0.682  &  0.234 & 17.82 & 0.547  &  \underline{0.420}  &  15.80 &  0.153 &   0.838\\
                Flower  & 14.58  & 0.423  & 0.526 & 12.91  & 0.415  & 0.589  & \underline{22.27}  &  \underline{0.683} & \underline{0.230} & 17.86  & \underline{0.549}  &  \underline{0.417}  & \underline{24.88}   & \underline{0.742}   &  \underline{0.247}\\
                D-Flow & \underline{19.96} & \underline{0.463}  &  \underline{0.312}  &  13.13 & 0.238  &  0.765 &  11.81 &  0.150 &  1.139  &   8.54 &  0.077 & 1.204  & 9.98 & 0.041 & 1.201\\
                P-Flow   & \textbf{20.34}  &  \textbf{0.620} & \textbf{0.262} &  \textbf{17.52} &  \textbf{0.527} & \textbf{0.341}  &  \textbf{24.66} &  \textbf{0.760} & \textbf{0.154} & \textbf{20.97}  & \textbf{0.640}  &  \textbf{0.254} & \textbf{25.50} &  \textbf{0.749} & \textbf{0.180}\\
                \bottomrule
            \end{tabular}
        }
    \end{subtable}

    \vspace{1.2em}

    \begin{subtable}{1\textwidth}
        \centering
        \caption{AFHQ-Cat dataset}
        \resizebox{\textwidth}{!}{
            \begin{tabular}{lccccccccccccccc}
                \toprule
                \multirow{2.5}{*}{Method}
                  & \multicolumn{3}{c}{$8\times\text{SR}$}
                  & \multicolumn{3}{c}{$16\times\text{SR}$}
                  & \multicolumn{3}{c}{$95\%$ Random inp.}
                  & \multicolumn{3}{c}{$99\%$ Random inp.} 
                  & \multicolumn{3}{c}{$\sigma=1.0$ Denoising} \\
                \cmidrule(lr){2-4}\cmidrule(lr){5-7}\cmidrule(lr){8-10}\cmidrule(lr){11-13}\cmidrule(lr){14-16}
                & PSNR  & SSIM  & LPIPS  & PSNR  & SSIM  & LPIPS  & PSNR  & SSIM  & LPIPS  & PSNR  & SSIM  & LPIPS  & PSNR  & SSIM  & LPIPS \\
                \midrule 
                Degraded &  11.86 & 0.317  &  0.958 &  11.87 & 0.353  &  0.885 &  12.11 & 0.252 & 1.119 & 12.02  & 0.302  & 1.039 & 9.15  &  0.037 &   1.314   \\
                Flow-Priors &  \underline{18.46}  &  0.371  &   \underline{0.482} & \underline{15.12}   & 0.301   &  0.611  & 21.74 &  0.532  & 0.400 &  \underline{21.27} &  0.516  &  0.427 & 17.23   &   0.173 &  0.795 \\
                PnP-Flow & 16.70  & 0.448  & 0.517 &  14.68 & 0.419  &  0.592 & 23.38  & 0.665  & 0.239  & 19.06  &  \underline{0.540} &   0.420  & 16.19 & 0.166 &  0.773  \\
                Flower  &  16.60  &  \underline{0.475} & 0.506 &  14.63 & \underline{0.437}  &  \underline{0.588} &  23.44 &  0.666  &  0.236 &  19.10 & \underline{0.540}  &  \underline{0.417} & \underline{24.89} & \underline{0.693} & \underline{0.339}  \\
                D-Flow &  12.56  &  0.125 &  1.054  &  9.62  &  0.079  &  1.109  &   \underline{25.86} &  \underline{0.710} &   \textbf{0.126} &  19.52  &  0.438  & 0.519   &  10.56  &  0.050  &  1.142 \\
                P-Flow   &  \textbf{23.32} & \textbf{0.621}  &  \textbf{0.254} &  \textbf{19.70} &  \textbf{0.524} &  \textbf{0.341} &  \textbf{26.38} &  \textbf{0.731} &  \underline{0.154} & \textbf{22.86}  & \textbf{0.626}  &  \textbf{0.236}  & \textbf{25.73} & \textbf{0.717}  & \textbf{0.207}\\
                \bottomrule
            \end{tabular}
        }
    \end{subtable}
\end{table}

\subsection{With high measurement noise}\label{app:With high measurement noise}

\begin{figure}
    \centering
    \includegraphics[width=1.0\linewidth]{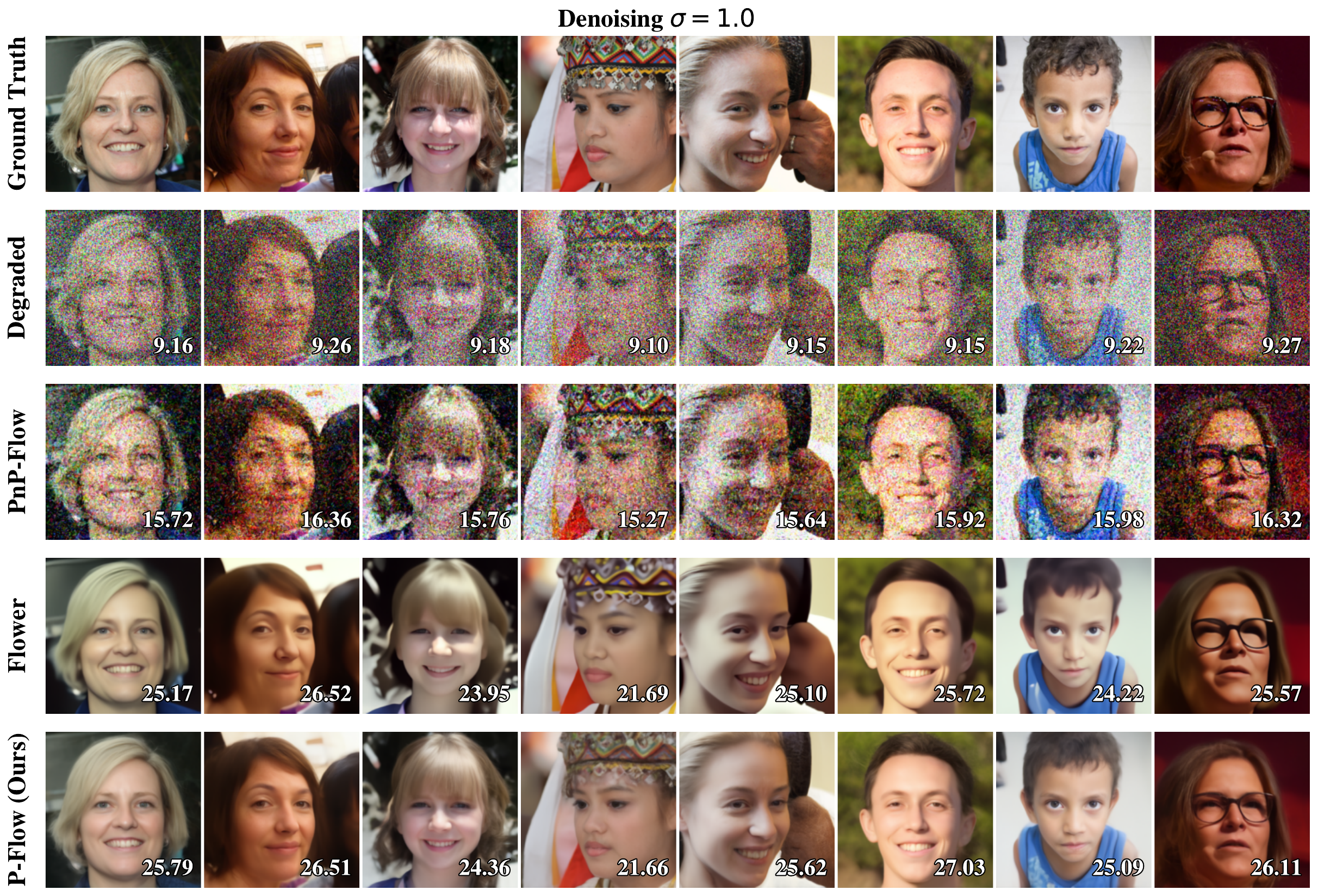}
    \caption{Visual comparison for denoising $\sigma=1.0$ on the FFHQ dataset. We compare the proposed P-Flow with PnP-Flow and Flower. The PSNR metric for each restoration is indicated in the bottom-right corner.}
    \label{fig:denoising sigma1.0}
\end{figure}

Under the high measurement noise condition (Figure~\ref{fig:denoising sigma1.0}), both P-Flow and Flower remain structurally stable, whereas PnP-Flow exhibits severe visual distortion. Flower avoids complete collapse, but it introduces a noticeable color shift and produces smoother, less detailed reconstructions. In contrast, P-Flow preserves sharper textures and achieves higher visual fidelity.
\par
This difference can be attributed to how each method incorporates unreliable measurement information. PnP-Flow applies data-fidelity updates directly to the intermediate iterates before reprojecting them onto the flow path and applying the time-dependent denoiser. When the observation is dominated by noise, these updates become highly unreliable and can move the intermediate states away from the time marginals for which the flow-based denoiser was trained. This makes the subsequent denoising step less effective and can lead to the catastrophic distortions observed in Figure~\ref{fig:denoising sigma1.0}.
\par
Flower is more robust because it refines the predicted destination through a measurement-aware proximal step rather than injecting a raw gradient into the intermediate state. However, when the measurement noise is extremely large, the refinement signal becomes weak and ambiguous. In this regime, Flower tends to behave like a conservative destination-refinement method: it preserves global structure, but may suppress high-frequency details and bias low-frequency image statistics, which manifests as over-smoothing and color shift.
\par
P-Flow differs structurally from both baselines. It does not modify the learned vector field or interrupt the intermediate flow trajectory. Instead, the noisy measurement is used only to guide the update of the latent source variable. The Gaussian spherical projection further keeps the optimized source within the typical region of the prior. As a result, P-Flow uses the noisy observation to select among plausible generative trajectories, while the final image is still produced by an unmodified source-to-data flow. This separation between measurement guidance and generative dynamics makes P-Flow more robust when the data-fidelity signal is heavily corrupted.

\end{document}